\pdfoutput=1

\documentclass[11pt]{article}

\usepackage[preprint]{acl}
\usepackage{soul}
\usepackage{times}
\usepackage{latexsym}
\usepackage{pifont}
\usepackage{enumitem}
\usepackage[T1]{fontenc}

\usepackage[utf8]{inputenc}

\usepackage{microtype}

\usepackage{inconsolata}

\usepackage{graphicx}
\usepackage{multirow}
\usepackage{amsmath,amssymb,amsthm}

\newtheorem{theorem}{Theorem}

\newtheorem{definition}{Definition}
\newtheorem{assumption}{Assumption}
\newtheorem{lemma}{Lemma}
\usepackage{ragged2e}
\usepackage{hyperref}       
\usepackage{url}            
\usepackage{colortbl}
\usepackage{subfigure}
\usepackage{array}
\usepackage{tabularx}
\usepackage{makecell}

\newcolumntype{C}[1]{>{\Centering\arraybackslash}p{#1}}

\usepackage{booktabs}
\usepackage{xcolor}
\definecolor{mycolor}{rgb}{0.5,0.1,0.8} 
\usepackage{xspace}
\usepackage{xcolor}
\usepackage{tikz}
\usepackage{titletoc}
\usepackage{svg}
\usepackage{amsmath}
\usepackage{soul}
\usepackage{wrapfig}

\newcommand{\algname}{DART\xspace}
\newcommand{\secref}[1]{\S\ref{#1}}
\newcommand{\mypara}[1]{\smallskip\noindent\textbf{#1}}

\newcommand{\fg}[1]{\mathbf{\mathcolor{ForestGreen}{#1}}}

\title{Stop Looking for ``Important Tokens'' in Multimodal Language Models:  \\
Duplication Matters More}

\author{
    {\bf Zichen Wen}$^{1, 2}$ \quad 
    {\bf Yifeng Gao}$^1$ \quad
    {\bf Shaobo Wang}$^{1}$ \quad
    {\bf Junyuan Zhang}$^{2}$ \quad
    {\bf Qintong Zhang}$^{2, 4}$ \\
    {\bf Weijia Li}$^{3, 2}$ \quad
    {\bf Conghui He}$^{2}$\footnotemark[2] \quad
    {\bf Linfeng Zhang}$^{1}$\footnotemark[2] \\
    \textsuperscript{1}Shanghai Jiao Tong University 
    \quad
    \textsuperscript{2}Shanghai AI Laboratory  \\
    \textsuperscript{3}Sun Yat-sen University \quad 
    \textsuperscript{4}Peking University \\
    \normalsize
    \texttt{zichen.wen@outlook.com, heconghui@pjlab.org.cn, zhanglinfeng@sjtu.edu.cn}
}

\begin{document}
\maketitle
{
\renewcommand{\thefootnote}{\fnsymbol{footnote}}
\footnotetext[2]{Corresponding authors.}
}

\begin{abstract}
Vision tokens in multimodal large language models often dominate huge computational overhead due to their excessive length compared to linguistic modality.
Abundant recent methods aim to solve this problem with token pruning, which first defines an importance criterion for tokens and then prunes the unimportant vision tokens during inference. 
However, in this paper, we show that the importance is not an ideal indicator to decide whether a token should be pruned. Surprisingly, it usually results in inferior performance than random token pruning and leading to incompatibility to efficient attention computation operators.
Instead, we propose \textbf{DART} (\textbf{D}uplication-\textbf{A}ware \textbf{R}eduction of \textbf{T}okens), which prunes tokens based on its duplication with other tokens, leading to significant and training-free acceleration.
Concretely, DART selects a small subset of pivot tokens and then retains the tokens with low duplication to the pivots, ensuring minimal information loss during token pruning. 
Experiments demonstrate that DART can prune \textbf{88.9\%} vision tokens while maintaining comparable performance, leading to a \textbf{1.99$\times$} and \textbf{2.99$\times$} speed-up in total time and prefilling stage, respectively, with good compatibility to efficient attention operators~\footnote{Our code is available at \url{https://github.com/ZichenWen1/DART}}. 

\end{abstract}

\section{Introduction}\label{sec:introduction}
\newcommand{\highlightedtextred}[1]{%
  \sethlcolor{red!20}%
  \hl{#1}%
}

\newcommand{\highlightedtextblue}[1]{%
  \sethlcolor{cyan!20}%
  \hl{#1}%
}

\definecolor{textred}{RGB}{255,0,0}

\definecolor{textgreen}{RGB}{112,173,71}

\definecolor{textblue}{RGB}{46,117,182}

\begin{figure}[t]
    \centering
    \includegraphics[width=1.02\linewidth]{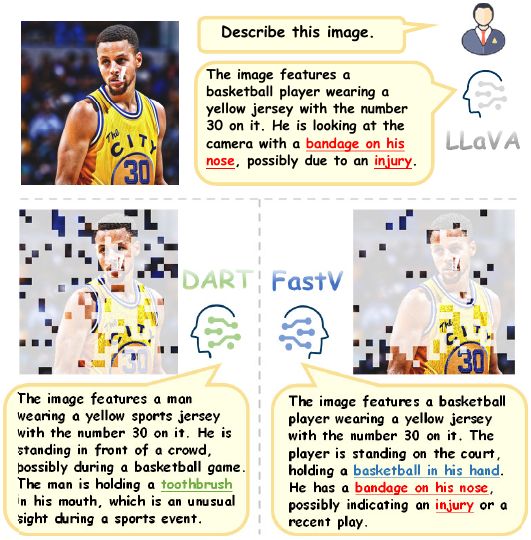}
    \vspace{-6mm}
    \caption{\textbf{Comparison between \algname and FastV.} \textcolor{textred}{\ul{Red text}} indicates hallucination from vanilla LLaVA-1.5-7B, \textcolor{textgreen}{\ul{green text}} represents hallucination from DART, and \textcolor{textblue}{\ul{blue text}} represents hallucination from FastV.}
    \vspace{-6mm}
    \label{fig:teaser_curry}
\end{figure}


Multimodal large language models (MLLMs) exhibit remarkable capabilities across a diverse range of multimodal tasks, including image captioning, visual question answering (VQA), video understanding~\citep{wang2024internvideo2}, and multimodal reasoning~\citep{wang2024exploring,kang2025legion}. 
\begin{figure}[t]
    \centering
    \includegraphics[width=1.0\linewidth]{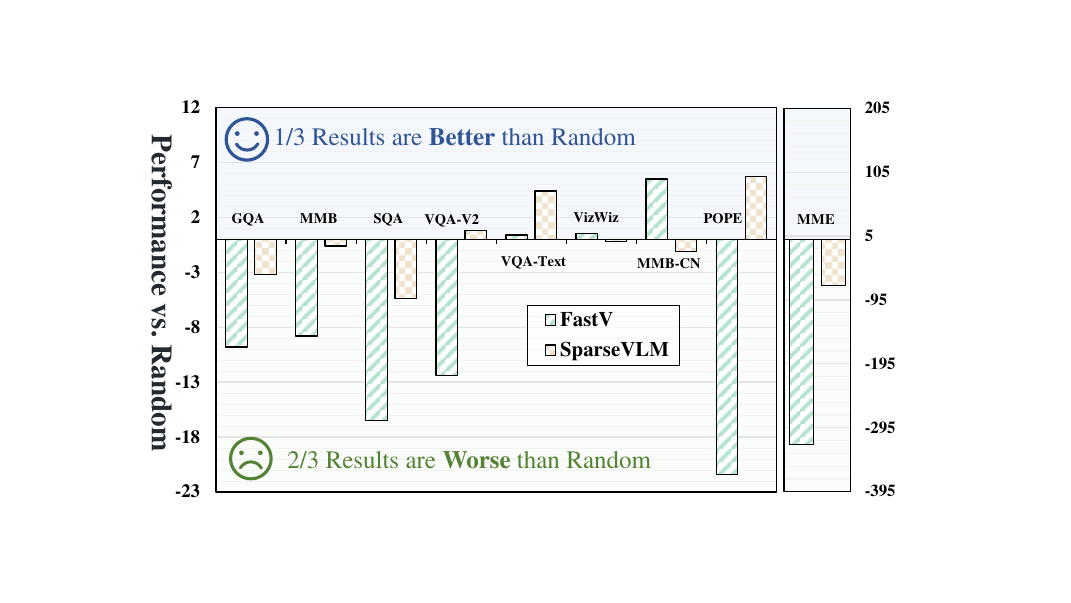}
    \vspace{-6mm}
    \caption{\textbf{Performance of FastV and SparseVLM compared with random token pruning} on the LLaVA-1.5-7B, with a \textbf{88.9\%} token reduction ratio.}
    \vspace{-7mm}
    \label{fig:random_vs_others}
\end{figure}
 However, such impressive performance is always accompanied by huge computation costs, which are mainly caused by massive vision tokens in the input data, especially for high-resolution images~\citep{li2024mini} and multi-frame video~\citep{tang2023video}, leading to challenges in their applications.

To solve this problem, abundant recent methods introduce \emph{token pruning} to remove the vision tokens in a training-free manner, which usually first defines the importance score of each token, and then prunes the most unimportant tokens during the inference phrase~\citep{chen2024image, zhang2024sparsevlm, liu2024multi}.  The key to a token pruning method is the definition of the importance of vision tokens, where most existing methods are based on the attention scores between vision-only tokens and vision-language tokens. However, this paper argues that these importance-based methods have several serious problems.

\noindent \textbf{(I) Ignoring interactions between tokens during pruning:} Although the interaction between different tokens is considered in attention scores, however, importance-based methods directly remove the most unimportant tokens, ignoring the truth that the importance of each token should be adjusted when other tokens are pruned or preserved. For instance, for two similar tokens, if one of both is determined to be pruned, then the importance of the other token should be improved and vice versa. Unfortunately, previous importance-based token pruning methods fail to model such interaction.

\noindent \textbf{(II) Incompatibility to efficient attention:} Efficient attention operators such as FlashAttention~\cite{dao2022flashattention} have become the default configure in neural networks, which accelerates attention computation by around 2$\times$ and reduce the memory costs from $O(N^2)$ to $O(N)$. 
However, these efficient attention operators make attention scores not accessible during computation, indicating conflicts with most previous importance-based token pruning methods. Disabling FlashAttention for accessing attention scores significantly improves the overall latency and memory footprint.

\noindent \textbf{(III) Bias in token positions:} As claimed by abundant recent works~\citep{endo2024feather, zhang2024cls} and shown in Figure~\ref{fig:teaser_curry}, attention scores have position bias, where the tokens are positionally close to the last token tend to have a higher attention score, making attention score does not truly reveal the value of this token.

\noindent \textbf{(IV) Significant accuracy drop:} Although the aforementioned three problems have reminded us of the ineffectiveness of importance-based token pruning, however, it is still extremely surprising to find that \textbf{\emph{{some influential importance-based token pruning methods show inferior accuracy than random token pruning}}}, (\emph{i.e.}, randomly selecting the tokens for pruning), as shown in Figure~\ref{fig:random_vs_others}.

The above observations demonstrates the disadvantages of importance-based token pruning methods, while also introducing the expectation for the ideal alternative: The expected method should consider both the individual value of a token and its interaction to other tokens. It should be cheap in computation and friendly to hardware, and shows no bias in the positions of tokens.


These insights inspire us to incorporate token duplication into the token reduction. Intuitively, when multiple tokens exhibit identical or highly similar representations, it is natural to retain only one of them for the following computation, thereby maintaining efficiency without harming accuracy. Building upon this idea, we introduce a simple but effective token pruning pipeline referred to as \textbf{DART} (\textbf{D}uplication-\textbf{A}ware
\textbf{R}eduction of \textbf{T}okens) with the following two steps.
\begin{figure*}[!h]
    \vspace{-2mm}
    \centering
    \includegraphics[width=\linewidth]{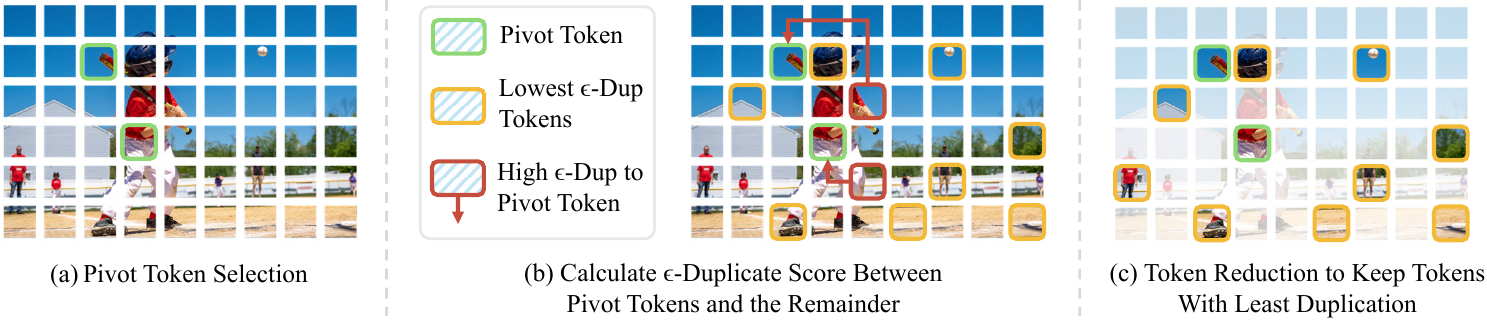}
    \vspace{-6mm}
    \caption{\textbf{The overview of DART.} The process includes (a) selecting pivot tokens, (b) calculating $\epsilon$-Duplicate scores between pivot tokens and other tokens, and (c) reducing tokens to retain those with the least duplication.}
    \vspace{-6mm}
    \label{fig:overview}
\end{figure*}

Firstly, we begin by selecting a small subset of tokens as pivot tokens, which comprise no more than $2\%$ of the total tokens. Such pivot tokens can be selected based on the norm of tokens or even randomly selected, which does not introduce notable computations.
Secondly, we then calculate the cosine similarity between pivot tokens and the remaining image tokens. Since the pivot tokens are fewer than $2\%$, such computation is efficient in both computing and memory.
With a desired token reduction ratio, we retain only those vision tokens with the lowest cosine similarity to pivot tokens and remove the similar ones.
The entire process is simple and highly efficient, completing in no more than \textbf{0.08} seconds, friendly to efficient attention operators, and leading to significantly higher accuracy than previous methods.

In summary, our contributions  are three-fold:
\begin{itemize}[leftmargin=10pt, topsep=0pt, itemsep=1pt, partopsep=1pt, parsep=1pt]
    \item \textbf{Rethink Token Importance.} Through empirical analysis, we demonstrate the suboptimality of relying on attention scores to measure token importance to guide the token reduction paradigm.
    \item \textbf{Token Duplication as a Key Factor.} Building on token duplication, we introduce a training-free, plug-and-play token reduction method that seamlessly integrates with Flash Attention.
    \item \textbf{Superior Performance with Extreme Compression.} Extensive experiments across four diverse MLLMs and over 10 benchmarks demonstrate the clear superiority of \algname. For instance, our method outperforms the second-best method by 2.2\% ($93.7\% \text{ vs. } 91.5\%$) on LLaVA-1.5-7B with an 88.9\% reduction ratio.
\end{itemize}

\vspace{-2mm}
\section{Related Work}
\vspace{-1mm}
\noindent \textbf{Multimodal Large Language Models}
Multimodal large language models (MLLMs) \citep{liu2024improved, li2023blip, zhu2023minigpt, liu2024visual} excel at image, video, and multimodal reasoning by integrating vision and text~\citep{zhang2024ocr}. However, visual data processing is costly due to redundancy, low information density \citep{liang2022evit,liu2025shifting}, and the quadratic cost of attention \citep{vaswani2017attention}. For instance, models like LLaVA \citep{liu2023improvedllava} and mini-Gemini-HD \citep{li2024mini} encode high-resolution images into thousands of tokens, while video models like VideoLLaVA \citep{lin2023video} and VideoPoet \citep{kondratyuk2023videopoet} handle even more tokens across frames. These challenges highlight the need for efficient token representations and longer context. Recent work like Gemini \citep{geminiteam2023gemini} and LWM \citep{liu2024world} addresses this by improving token efficiency and extending context, enabling more scalable MLLMs.

\noindent \textbf{Visual Token Compression}
Visual tokens often outnumber text tokens by tens to hundreds of times, as visual signals are more spatially redundant than information-dense text \citep{marr2010vision}. 
LLaMA-VID \citep{li2023llama} employs a Q-Former with context tokens, and DeCo \citep{yao2024deco} uses adaptive pooling.
DTMFormer~\citep{wang2024dtmformer} improves ViTs' efficiency in medical image segmentation by merging redundant tokens during training.
MADTP~\citep{cao2024madtp} reduces computation by aligning cross-modal features and pruning tokens.
However, these require modifying components and additional training.
ToMe~\citep{bolya2022tome} merges tokens without training but disrupts cross-modal interactions~\citep{xing2024PyramidDrop}. FastV~\citep{chen2024image} selects via attention scores, while SparseVLM~\citep{zhang2024sparsevlm} uses text guidance. 
Yet, these forgo Flash-Attention~\citep{dao2022flashattention, dao2023flashattention2}, neglecting token duplication.
We preserve hardware acceleration (\emph{i.e.}, Flash-Attention) and target duplication for efficient token reduction.

\vspace{-1mm}
\section{Methodology}
\vspace{-1mm}
\subsection{Preliminary}
\label{sec:preliminary}
\vspace{-2mm}
\mypara{Architecture of MLLM.} The architecture of Multimodal Large Language Models (MLLMs) typically comprises three core components: a visual encoder, a modality projector, and a language model (LLM). Given an image $I$, the visual encoder and a subsequent learnable MLP are used to encode $I$ into a set of visual tokens $e_v$. These visual tokens $e_v$ are then concatenated with text tokens $e_t$ encoded from the text prompt $p_t$, forming the input for the LLM. The LLM decodes the output tokens $y$ sequentially, which can be formulated as:
    $y_i = f(I, p_t, y_0, y_1, \cdots, y_{i-1}).$


\vspace{-1mm}
\subsection{Beyond Token Importance: Questioning the Status Quo}
Given the computational burden associated with the length of visual tokens in MLLMs, numerous studies have embraced a paradigm that utilizes attention scores to evaluate the significance of visual tokens, thereby facilitating token reduction.
Specifically, in transformer-based MLLMs, each layer performs attention computation as illustrated below:
\begin{equation}
   \text{Attention}(\mathbf{Q}, \mathbf{K}, \mathbf{V}) = \text{softmax}\left(\frac{\mathbf{Q} \cdot \mathbf{K}^\top}{\sqrt{d_k}}\right)\cdot \mathbf{V},
\end{equation}
where $d_k$ is the dimension of $\mathbf{K}$. The result of $\text{Softmax}(\mathbf{Q}\cdot \mathbf{K}^\top/\sqrt{d_k})$ is a square matrix known as the attention map.
Existing methods extract the corresponding attention maps from one or multiple layers and compute the average attention score for each visual token based on these attention maps:
\begin{equation}
    \phi_{\text{attn}}(x_i) = \frac{1}{N} \sum_{j=1}^{N} \text{Attention}(x_i, x_j),
\end{equation}
where $\text{Attention}(x_i, x_j)$ denotes the attention score between token $x_i$ and token $x_j$, $\phi_{\text{attn}}(x_i)$ is regarded as the importance score of the token $x_i$, $N$ represents the number of visual tokens.
Finally, based on the importance score of each token and the predefined reduction ratio, the most important visual tokens are selectively retained:
\begin{equation}
    \mathcal{R} = \{ x_i \mid (\phi_{\text{attn}}(x_i) \geq \tau) \},
\end{equation}
where $\mathcal{R}$ represents the set of retained visual tokens, and $\tau$ is a threshold determined by the predefined reduction ratio.

\noindent{\textbf{Problems:}} Although this paradigm has demonstrated initial success in enhancing the efficiency of MLLMs, it is accompanied by several inherent limitations that are challenging to overcome.

One key limitation is disregarding the dynamic nature of token importance during pruning.
For a token sequence \(\{x_1, \ldots, x_n\}\), importance-based methods compute static token importance via a scoring function \(s_i = \mathcal{F}(x_i | X)\), where \(X\) is the full token set. The strategy retains Top-\(k\) tokens:
\begin{equation}
    X_{\text{pruned}} = \arg \max_{X' \subseteq X, |X'| = k} \sum_{x_j \in X'} s_j
\end{equation}
This implies an \textbf{independence assumption}: the score \( s_j \) remains unchanged for any subset \( X' \subset X \), ignoring dynamic token interactions. For example, if two similar tokens \( x_p, x_q \) have \( s_p \approx s_q \), removing \( x_q \) should recalibrate \( s_p \) as:
\begin{equation}
    s_p' = \mathcal{F}(x_p | X' \setminus \{x_q\}) > s_p,
\end{equation}
which leads to a bias in importance estimation \(\Delta = s_p' - s_p\). This contradiction between static scoring and dynamic interaction can be quantified as:
\begin{equation}
    \mathbb{E}_{X' \subset X} \left[ \sum_{x_i \in X'} \left( \mathcal{F}(x_i | X') - \mathcal{F}(x_i | X) \right) \right]
\end{equation}


\vspace{-2mm}
Additionally, Figure~\ref{fig:teaser_curry} visualizes the results of token reduction, revealing that selecting visual tokens based on attention scores introduces a noticeable bias toward tokens in the lower-right region of the image, those appearing later in the visual token sequence. However, this region is not always the most significant in every image. Further, we present the outputs of various methods. Notably, FastV generates more hallucinations than the vanilla model, while \algname effectively reduces them. 
We attribute this to the inherent bias of attention-based methods, which tend to retain tokens concentrated in specific regions, often neglecting the broader context of the image. In contrast, \algname removes highly duplication tokens and preserves a more balanced distribution across the image, enabling more accurate and consistent outputs.

Furthermore, methods relying on attention scores for token importance are incompatible with Flash Attention, compromising speed, and sometimes even underperforming random token reduction in effectiveness (See Fig.~\ref{fig:random_vs_others}).


\begin{figure*}[!ht]
    \centering
    \includegraphics[width=1.02\linewidth]{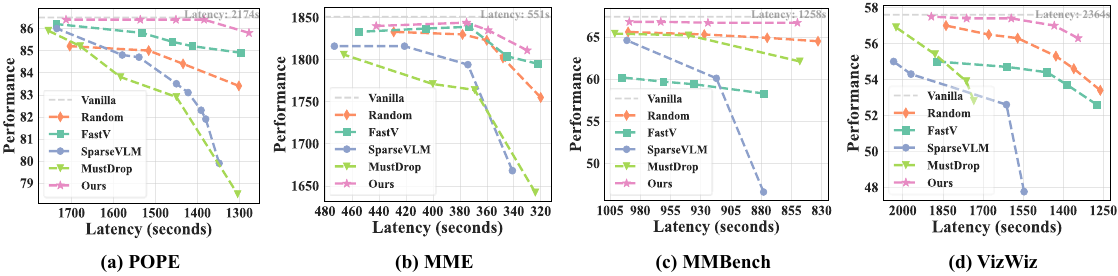}
    \vspace{-7mm}
    \caption{\textbf{Performance-Latency trade-off comparisons} across different datasets on LLaVA-Next-7B. \algname consistently achieves better performance under varying latency constraints compared to other approaches.}
    \vspace{-4mm}
    \label{fig:latency_vs_performance}
\end{figure*}

\subsection{Token Duplication: Rethinking Reduction}
Given the numerous drawbacks associated with the paradigm of using attention scores to evaluate token importance for token reduction, \textit{what additional factors should we consider beyond token importance in the process of token reduction?}
Inspired by the intuitive ideas mentioned in \secref{sec:introduction} and the phenomenon of tokens in transformers tending toward uniformity (\emph{i.e.}, over-smoothing)~\citep{nguyen2023mitigating, gong2021vision}, we propose that token duplication should be a critical focus.

Due to the prohibitively high computational cost of directly measuring duplication among all tokens, we adopt a paradigm that involves selecting a minimal number of pivot tokens.
\begin{definition}[Pivot Tokens]
Let $ \mathcal{P} = \{p_1, p_2, \dots, p_k\} \subseteq \mathcal{X} $ denote the  pivot tokens, where $ k \ll n $ and $ n $ is the total length of the tokens $ \mathcal{X} = \{x_1, x_2, \dots, x_n\}$. The pivot tokens $\mathcal{P}$ are a subset of $ \mathcal{X} $, selected for their representativeness of the entire set.
\end{definition}

Given the pivot tokens, we can define the duplication score based on it.
\begin{definition}[$\epsilon$-duplicate Score]\label{def:dup}
The token duplication score between a pivot token $ p_i $ and a visual token $ x_j $ is defined as: 
\begin{equation}
    \text{dup}(p_i, x_j) = \frac{p_i^\top x_j}{\| p_i \|  \| x_j \|},
\end{equation}
where $ \| \cdot \| $ denotes the Euclidean norm. Two tokens $ p_i, x_j $ are \textbf{$ \epsilon $-duplicates} if
\begin{equation}
    \text{dup}(p_i, x_j)  > \epsilon.
\end{equation}
\end{definition}
With the $\epsilon$-duplicate score, for each pivot \( p_i \), the associated retained token set is defined as:
\begin{equation}
    \mathcal{R}_i = \{ x_j \mid dup(p_i, x_j) \leq \epsilon \}
\end{equation}
The final retained set is:
\begin{equation}
    \mathcal{R} = \mathcal{P} \cup \left( \bigcup_{p_i \in \mathcal{P}} \mathcal{R}_i \right)
\end{equation}
where \( \epsilon \) is the threshold dynamically determined for each pivot \( p_i \) based on reduction ratio.
This ensures that only tokens that are sufficiently different from the pivot tokens are kept.


Our method is orthogonal to the paradigm of using attention scores to measure token importance, meaning it is compatible with existing approaches. Specifically, we can leverage attention scores to select pivot tokens, and subsequently incorporate token duplication into the process.

However, this still does not fully achieve compatibility with Flash Attention. Therefore, we explored alternative strategies for selecting pivot tokens, such as using K-norm, V-norm\footnote{Here, the K-norm and V-norm refer to the L1-norm of K matrix and V matrix in attention computing, respectively.}, or even random selection. Surprisingly, all these strategies achieve competitive performance across multiple benchmarks. This indicates that our token reduction paradigm based on token duplication is not highly sensitive to the choice of pivot tokens. Moreover, it suggests that removing duplicate tokens may be more critical than identifying ``important tokens'', highlighting token duplication as a more significant factor in token reduction.
Detailed discussion on pivot token selection is provided in \secref{pivot_token_selection}.

\subsection{Theoretical Analysis}
To further justify trustworthiness of our proposed method, we provide a theoretical analysis of it.
\begin{assumption}[Transformer Property]\label{assump:transformer} For transformer property, we assume the following: \\
\noindent \textbf{(A1)}. (Lipschitz continuity under Hausdorff distance). The model $f$ is Lipschitz continuous with respect to the Hausdorff distance between token sets. Formally, there exists $K > 0$ such that for any two token sets $\mathcal{X}_1, \mathcal{X}_2 \subseteq \mathbb{R}^d$:
    $$
    \|f(\mathcal{X}_1) - f(\mathcal{X}_2)\| \leq K \cdot d_H(\mathcal{X}_1, \mathcal{X}_2),
    $$
    where $d_H(\mathcal{X}_1, \mathcal{X}_2) \triangleq \max$\\ $\left\{\sup\limits_{x_1 \in \mathcal{X}_1} \inf\limits_{x_2 \in \mathcal{X}_2} \|x_1 - x_2\|, \sup\limits_{x_2 \in \mathcal{X}_2} \inf\limits_{x_1 \in \mathcal{X}_1} \|x_1 - x_2\|\right\}$.
\noindent \textbf{(A2)}. (Bounded embedding). All tokens have bounded Euclidean norms:
    $$
    \|x\| \leq B, \quad \forall x \in \mathcal{X},
    $$
    where $B > 0$ is a constant.
\end{assumption}
\begin{lemma}[Bounded Distance]\label{lemma:cosine} $\min_{p_i\in \mathcal{P}} |p_i-x_j|\leq (2(1-\epsilon))^{1/2}B,\quad \forall x_j\in \mathcal{X}\setminus \mathcal{R}$.
\end{lemma}
\begin{proof} Using A2 and Definition~\ref{def:dup}, we obtain:
\begin{equation*}
    \begin{aligned}
        & \min_{p_i\in \mathcal{P}} |p_i-x_j|^2  = \min_{p_i\in \mathcal{P}} (|p_i|^2 + |x_j|^2 - 2 p_i^\top x_j) \\
        & \leq \min_{p_i\in \mathcal{P}} (B^2 + B^2 - 2 \epsilon \cdot B \cdot B) \leq 2(1-\epsilon) B^2
    \end{aligned}
\end{equation*}
Therefore, the duplication distance bound is given by:
$\min_{p_i\in \mathcal{P}} |p_i-x_j|^2\leq (2(1-\epsilon))^{1/2}B$
\end{proof}
\vspace{-2mm}
\begin{lemma}[Bounded Approximation Error]\label{lemma:bound}
Under Assumption~\ref{assump:transformer}, the Hausdorff distance between original and retained tokens satisfies:
$$
d_H(\mathcal{X}, \mathcal{R}) \leq \sqrt{2(1-\epsilon)}B.
$$
\end{lemma}
\begin{proof}
For any $x \in \mathcal{X}$:
\begin{itemize}[leftmargin=10pt, topsep=0pt, itemsep=1pt, partopsep=1pt, parsep=1pt]
    \item If $x \in \mathcal{R}$, then $\inf_{r \in \mathcal{R}} \|x - r\| = 0$
    \item If $x \notin \mathcal{R}$, by definition and Lemma~\ref{lemma:cosine} there exists $p_i \in \mathcal{P} \subseteq \mathcal{R}$ with $\|x - p_i\| \leq \sqrt{2(1-\epsilon)}B$
\end{itemize}
Thus:
$$
\sup_{x \in \mathcal{X}} \inf_{r \in \mathcal{R}} \|x - r\| \leq \sqrt{2(1-\epsilon)}B.
$$
Since $\mathcal{R} \subseteq \mathcal{X}$, Hausdorff distance simplifies to:
$d_H(\mathcal{X}, \mathcal{R}) = \sup_{x \in \mathcal{X}} \inf_{r \in \mathcal{R}} \|x - r\| \leq \sqrt{2(1-\epsilon)}B.$
\end{proof}
\begin{theorem}[Performance Guarantee]\label{thm:main}
Under Assumptions~\ref{assump:transformer}, the output difference between original and pruned token sets is bounded by:
$$
\|f(\mathcal{X}) - f(\mathcal{R})\| \leq K\sqrt{2(1-\epsilon)}B.
$$
\end{theorem}
\begin{proof}
Direct application of Lipschitz continuity (A1) with Lemma~\ref{lemma:bound}: $\|f(\mathcal{X}) - f(\mathcal{R})\| \leq K \cdot d_H(\mathcal{X}, \mathcal{R}) \leq K\sqrt{2(1-\epsilon)}B.$
\end{proof}
This provides a theoretical guarantee that \algname preserves model output within a controllable bound, thereby supporting the trustworthiness and robustness of our method.

\renewcommand{\multirowsetup}{\centering}
\definecolor{mygray}{gray}{.92}
\definecolor{ForestGreen}{RGB}{34,139,34}
\definecolor{Forestred}{RGB}{220,50,50}
\begin{table*}[!ht]
    \centering
    \vspace{-1mm}
    \setlength{\tabcolsep}{3.5pt}
    \renewcommand{\arraystretch}{0.9}
    \footnotesize
    \centering
    \scalebox{0.9}{
    \begin{tabular}{c | c c c c c c c c c c| >{\centering\arraybackslash}p{1.0cm}}
        \toprule[1.5pt]
        \textbf{Method} & \textbf{GQA} & \textbf{MMB} & \textbf{MMB-CN} & \textbf{MME} & \textbf{POPE} & \textbf{SQA} & \textbf{VQA}$^{\text{V2}}$ & \textbf{VQA}$^{\text{Text}}$ & \textbf{VizWiz} & \textbf{OCRBench} & \makecell[c]{\textbf{Avg}.}\\
        \hline
        \rowcolor{mygray}
        LLaVA-1.5-7B & \multicolumn{11}{c}{\textit{Upper Bound, 576 Tokens} \ $\textbf{(100\%)}$}\\
        \textcolor{gray}{Vanilla} & \textcolor{gray}{61.9} & \textcolor{gray}{64.7} & \textcolor{gray}{58.1} & \textcolor{gray}{1862} & \textcolor{gray}{85.9} & \textcolor{gray}{69.5} & \textcolor{gray}{78.5} & \textcolor{gray}{58.2} & \textcolor{gray}{50.0} & \textcolor{gray}{297} & \multirow{1}*{\textcolor{gray}{100.0\%}} \\
        \hline

        \rowcolor{mygray}
        LLaVA-1.5-7B & \multicolumn{11}{c}{\textit{Retain 192 Tokens} \ $\fg{(\downarrow 66.7\%)}$}\\
        ToMe \texttt{\scriptsize{(ICLR23)}} & 54.3 & 60.5 & - & 1563 & 72.4 & 65.2 & 68.0 & 52.1 & - & - & - \\
        FastV \texttt{\scriptsize{(ECCV24)}} & 52.7 & 61.2 & 57.0 & 1612 & 64.8 & 67.3 & 67.1 & 52.5 & 50.8 & 291 & \multirow{1}*{91.2\%} \\
        HiRED \texttt{\scriptsize{(AAAI25)}} & 58.7 & 62.8 & 54.7 & 1737 & 82.8 & 68.4 & 74.9 & 47.4 & 50.1 & 190 & 91.5\%     \\
        FitPrune \texttt{\scriptsize{(AAAI25)}} & \textbf{60.4 }& 63.3 & 56.4 & 1831 & 83.4 & 67.8 & - & 57.4 & 50.9 & - & - \\
        LLaVA-PruMerge \texttt{\scriptsize{(2024.05)}} & 54.3 & 59.6 & 52.9 & 1632 & 71.3 & 67.9 & 70.6 & 54.3 & 50.1 & 253 & 90.8\% \\
        \multirow{1}*{SparseVLM \texttt{\scriptsize{(ICML25)}}} & 57.6 & 62.5 & 53.7 & 1721 & \textbf{83.6} & 69.1 & 75.6 & 56.1 & 50.5 & 292 & 96.3\% \\
        PDrop \texttt{\scriptsize{(CVPR25)}} & 57.1 & 63.2 & 56.8 & 1766 & 82.3 & 68.8 & 75.1 & 56.1 & 51.1 & 290 & 96.7\% \\
        FiCoCo-V \texttt{\scriptsize{(2024.11)}} & 58.5 & 62.3 & 55.3 & 1732 & 82.5 & 67.8 & 74.4 & 55.7 & 51.0 & - & 96.1\% \\
        MustDrop \texttt{\scriptsize{(2024.11)}} & 58.2 & 62.3 & 55.8 & 1787 & 82.6 & 69.2 & 76.0 & 56.5 & \textbf{51.4} & 289 & 97.2\% \\
        \algname (Ours) & 60.0 & \textbf{63.6} & \textbf{57.0} & \textbf{1856} & 82.8 & \textbf{69.8} & \textbf{76.7} & \textbf{57.4} & 51.2 & \textbf{296} & \textbf{98.8\%} \\
        \rowcolor{cyan!7}
        \algname$^\dagger$ (Ours) & 60.9 & 66.3 & 59.5 & 1829 & 85.3 & 70.1 & 78.2 & 56.8 & 51.3 & 304 & 100.4\% \\
        \hline

        \rowcolor{mygray}
        LLaVA-1.5-7B & \multicolumn{11}{c}{\textit{Retain 128 Tokens} \ $\fg{(\downarrow 77.8\%)}$}\\
        ToMe \texttt{\scriptsize{(ICLR23)}} & 52.4 & 53.3 & - & 1343 & 62.8 & 59.6 & 63.0 & 49.1 & - & - & - \\
        FastV \texttt{\scriptsize{(ECCV24)}} & 49.6 & 56.1 & 56.4 & 1490 & 59.6 & 60.2 & 61.8 & 50.6 & 51.3 & 285 & \multirow{1}*{86.4\%}\\
        HiRED \texttt{\scriptsize{(AAAI25)}} & 57.2 & 61.5 & 53.6 & 1710 & 79.8 & 68.1 & 73.4 & 46.1 & 51.3 & 191 & 90.2\% \\
        FitPrune \texttt{\scriptsize{(AAAI25)}} & 58.5 & 62.7 & 56.2 & 1776 & 77.9 & 68.0 & - & 55.7 & 51.7 & - & - \\
        LLaVA-PruMerge \texttt{\scriptsize{(2024.05)}} & 53.3 & 58.1 & 51.7 & 1554 & 67.2 & 67.1 & 68.8 & 54.3 & 50.3 & 248 & 88.8\%  \\
        SparseVLM
         \texttt{\scriptsize{(ICML25)}} & 56.0 & 60.0 & 51.1 & 1696 & 80.5 & 67.1 & 73.8 & 54.9 & 51.4 & 280 & 93.8\% \\
        PDrop \texttt{\scriptsize{(CVPR25)}} & 56.0 & 61.1 & 56.6 & 1644 & \textbf{82.3} & 68.3 & 72.9 & 55.1 & 51.0 & 287 & 95.1\% \\
        FiCoCo-V \texttt{\scriptsize{(2024.11)}} & 57.6 & 61.1 & 54.3 & 1711 & 82.2 & 68.3 & 73.1 & 55.6 & 49.4 & - & 94.9\% \\
        MustDrop \texttt{\scriptsize{(2024.11)}} & 56.9 & 61.1 & 55.2 & 1745 & 78.7 & 68.5 & 74.6 & 56.3 & \textbf{52.1} & 281 & 95.6\% \\
        \algname (Ours) & \textbf{58.7} & \textbf{63.2} & \textbf{57.5} & \textbf{1840} & 80.1 & \textbf{69.1} & \textbf{75.9} & \textbf{56.4} & 51.7 & \textbf{296} & \textbf{98.0\%} \\
        \rowcolor{cyan!7}
        \algname$^\dagger$ (Ours) & 59.8 & 65.6 & 58.3 & 1849 & 84.4 & 70.7 & 77.5 & 56.4 & 52.6 & 299 & 99.9\% \\
        \hline

        \rowcolor{mygray}
        LLaVA-1.5-7B & \multicolumn{11}{c}{\textit{Retain 64 Tokens} \ $\fg{(\downarrow 88.9\%)}$}\\
        ToMe \texttt{\scriptsize{(ICLR23)}} & 48.6 & 43.7 & - & 1138 & 52.5 & 50.0 & 57.1 & 45.3 & - & - & - \\
        FastV \texttt{\scriptsize{(ECCV24)}} & 46.1 & 48.0 & 52.7 & 1256 & 48.0 & 51.1 & 55.0 & 47.8 & 50.8 & 245 & 77.3\% \\
        HiRED \texttt{\scriptsize{(AAAI25)}} & 54.6 & 60.2 & 51.4 & 1599 & 73.6 & 68.2 & 69.7 & 44.2 & 50.2 & 191 & 87.0\% \\
        FitPrune \texttt{\scriptsize{(AAAI25)}} & 52.3 & 58.5 & 49.7 & 1556 & 60.9 & 68.0 & - & 51.2 & 51.1 & - & - \\
        LLaVA-PruMerge \texttt{\scriptsize{(2024.05)}} & 51.9 & 55.3 & 49.1 & 1549 & 65.3 & 68.1 & 67.4 & 54.0 & 50.1 & 250 & 87.4\% \\
        SparseVLM \texttt{\scriptsize{(ICML25)}} & 52.7 & 56.2 & 46.1 & 1505 & 75.1 & 62.2 & 68.2 & 51.8 & 50.1 & 180 & 84.6\% \\
        PDrop \texttt{\scriptsize{(CVPR25)}} & 41.9 & 33.3 & 50.5 & 1092 & 55.9 & 68.6 & 69.2 & 45.9 & 50.7 & 250 & 78.1\% \\
        FiCoCo-V \texttt{\scriptsize{(2024.11)}} & 52.4 & 60.3 & 53.0 & 1591 & \textbf{76.0} & 68.1 & 71.3 & 53.6 & 49.8 & - & 91.5\% \\
        MustDrop \texttt{\scriptsize{(2024.11)}} & 53.1 & 60.0 & 53.1 & 1612 & 68.0 & 63.4 & 69.3 & 54.2 & 51.2 & 267 & 90.1\%    \\
        \algname (Ours) & \textbf{55.9} & \textbf{60.6} & \textbf{53.2} & \textbf{1765} & 73.9 & \textbf{69.8} & \textbf{72.4} & \textbf{54.4} & \textbf{51.6} & \textbf{270} & \textbf{93.7\%} \\
        \rowcolor{cyan!7}
        \algname$^\dagger$ (Ours) & 57.1 & 64.7 & 56.7 & 1823 & 79.3 & 71.1 & 74.6 & 54.7 & 52.1 & 286 & 97.2\% \\

        \bottomrule[1.5pt]
	\end{tabular}}
        \vspace{-2mm}
	\caption{Comparative experiments on image understanding. In all experiments for \algname, tokens are pruned after the second layer with $8$ pivot tokens. The pivot tokens are selected based on the maximum K-norm. \algname$^\dagger$ indicates that DART is applied during the training stage of LLaVA-1.5-7B.}
    \label{tab:main_1_5}
     \vspace{-2mm}
\end{table*}
\begin{table*}[!h]
    \centering
    \setlength{\tabcolsep}{2.0pt}
    \renewcommand{\arraystretch}{0.95}
    \footnotesize
    \scalebox{0.95}{
    \begin{tabular}{@{}lcccccccc}
        \toprule[1.2pt]
         \multirow{2}{*}{\textbf{Methods}} & \multirow{2}{*}{\textbf{Tokens $\downarrow$}} & \textbf{Total Time $\downarrow$} & \textbf{Prefilling Time $\downarrow$} & \multirow{2}{*}{\textbf{FLOPs $\downarrow$}} & \textbf{KV Cache $\downarrow$}  & \multirow{1}{*}{\textbf{POPE $\uparrow$}} & \multicolumn{2}{c}{\textbf{Speedup $\uparrow$}}   \\
           && \textbf{(Min:Sec)} & \textbf{(Min:Sec)} & & \textbf{(MB)} & \textbf{(F1-Score)} & \textbf{(Total)} & \textbf{(Prefilling)} \\
         \midrule
         \textcolor{gray}{Vanilla LLaVA-Next-7B} & \textcolor{gray}{2880} & \textcolor{gray}{36:16} & \textcolor{gray}{22:51} & \textcolor{gray}{100\%} & \textcolor{gray}{1512.1} &\textcolor{gray}{86.5} & \textcolor{gray}{1.00$\times$} & \textcolor{gray}{1.00$\times$} \\
         \hspace{0.5em} + FastV & 320 & 18:17 & 7:41 & \textbf{12.8\%} & 168.0 & 78.3 & 1.98$\times$ & 2.97$\times$ \\
         \hspace{0.5em} + SparseVLM & 320 & 23:11 & - & 15.6\% & 168.0 & 82.3 & 1.56$\times$ & - \\
          \rowcolor{cyan!10}
         \hspace{0.5em} + \algname & 320 &\textbf{18:13} & \textbf{7:38} & \textbf{12.8\%} & 168.0 & \textbf{84.1} & \textbf{1.99$\times$} & \textbf{2.99$\times$} \\
       
        \bottomrule[1.2pt]
    \end{tabular}}
    \vspace{-2mm}
    \caption{Inference costs of the number of tokens, Total-Time, Prefilling-Time, FLOPs, and KV Cache Memory.}
    \vspace{-6mm}
    \label{tab:efficiency}
\end{table*}
\section{Experiments}
\vspace{-0.1cm}
\noindent\textbf{Experiment Setting.} We conduct experiments on over four MLLMs across ten image-based and four video-based benchmarks. For details on implementation, please refer to Appendix~\ref{app:detailed_experiment_settings}.


\subsection{Main Results}
\renewcommand{\multirowsetup}{\centering}
\definecolor{mygray}{gray}{.92}
\definecolor{ForestGreen}{RGB}{34,139,34}
\definecolor{Forestred}{RGB}{220,50,50}
\begin{table*}[!ht]
    \centering
    \vspace{-1mm}
    \setlength{\tabcolsep}{3.5pt}
    \renewcommand{\arraystretch}{0.9}
    \footnotesize
    \centering
    \scalebox{0.9}{
    \begin{tabular}{c | c c c c c c c c c c| >{\centering\arraybackslash}p{1.0cm}}
        \toprule[1.5pt]
        \textbf{Method} & \textbf{GQA} & \textbf{MMB} & \textbf{MMB-CN} & \textbf{MME} & \textbf{POPE} & \textbf{SQA} & \textbf{VQA}$^{\text{V2}}$ & \textbf{VQA}$^{\text{Text}}$ & \textbf{VizWiz} & \textbf{OCRBench} & \makecell[c]{\textbf{Avg}.}\\
        \hline
        \rowcolor{mygray}
        LLaVA-Next-7B & \multicolumn{11}{c}{\textit{Upper Bound, 2880  Tokens} \ $\textbf{(100\%)}$}\\
         \textcolor{gray}{Vanilla} & \textcolor{gray}{64.2} & \textcolor{gray}{67.4} & \textcolor{gray}{60.6} & \textcolor{gray}{1851} & \textcolor{gray}{86.5} & \textcolor{gray}{70.1} & \textcolor{gray}{81.8} & \textcolor{gray}{64.9} & \textcolor{gray}{57.6} & \textcolor{gray}{517} &  \textcolor{gray}{100.0\%} \\
          \hline
       \rowcolor{mygray}
        LLaVA-Next-7B & \multicolumn{11}{c}{\textit{Retain 320 Tokens} \ $\fg{(\downarrow 88.9\%)}$} \\

        FastV \texttt{\scriptsize{(ECCV24)}} & 55.9 & 61.6 & 51.9 & 1661 & 71.7 & 62.8 & 71.9 & 55.7 & 53.1 & 374 & 86.4\% \\
        
        HiRED \texttt{\scriptsize{(AAAI25)}} & 59.3 & 64.2 & 55.9 & 1690 & 83.3 & 66.7 & 75.7 & 58.8 & 54.2 & 404 & 91.8\% \\
        
        LLaVA-PruMerge \texttt{\scriptsize{(2024.05)}} & 53.6 & 61.3 & 55.3 & 1534 & 60.8 & 66.4 & 69.7 & 50.6 & 54.0 & 146 & 79.9\% \\

       SparseVLM \texttt{\scriptsize{(ICML25)}} & 56.1 & 60.6 & 54.5 & 1533 & 82.4 & 66.1 & 71.5 & 58.4 & 52.0 & 270 & 85.9\%  \\

       PDrop \texttt{\scriptsize{(CVPR25)}} & 56.4 & 63.4 & 56.2 & 1663 & 77.6 & 67.5 & 73.5 & 54.4 & 54.1 & 259 & 86.8\% \\

        MustDrop \texttt{\scriptsize{(2024.11)}} & 57.3 & 62.8 & 55.1 & 1641 & 82.1 & 68.0 & 73.7 & \textbf{59.9} & 54.0 & 382 & 90.4\%    \\

       FasterVLM \texttt{\scriptsize{(2024.12)}} & 56.9 & 61.6 & 53.5 & 1701 & 83.6 & 66.5 & 74.0 & 56.5 & 52.6 & 401 & 89.8\% \\
       
       GlobalCom$^2$ \texttt{\scriptsize{(2025.01)}} & 57.1 & 61.8 & 53.4 & 1698 & 83.8 & 67.4 & 76.7 & 57.2 & 54.6 & 375 & 90.3\%  \\

       \algname (Ours)& \textbf{61.7} & \textbf{65.3} & \textbf{58.2} & \textbf{1710} & \textbf{84.1} & \textbf{68.4} & \textbf{79.1} & 58.7 & \textbf{56.1} & \textbf{406} & \textbf{93.9\%}  \\

        \bottomrule[1.5pt]
	\end{tabular}}
        \vspace{-2mm}
	\caption{Comparative experiments are performed on LLaVA-Next-7B using the same settings as LLaVA-1.5-7B.}
    \label{tab:main_1_6}
     \vspace{-6mm}
\end{table*}
\textbf{Image understanding task.}
\noindent The results presented in Tables~\ref{tab:main_1_5} and~\ref{tab:main_1_6} highlight \textbf{\algname}'s exceptional performance across diverse image understanding tasks under varying token configurations. 
We observe that \textbf{(i)} with only $192$ tokens, \algname achieves an impressive $98.8\%$ average performance, substantially outperforming second-best MustDrop by $\mathbf{1.6\%}$. 
\textbf{(ii)} This trend strengthens under aggressive reduction ratios, with \algname leading by $\mathbf{2.2\%}$ using just $64$ tokens.
\textbf{(iii)} Moreover, \algname scales seamlessly to advanced and larger models like LLaVA-Next-7B and Qwen2-VL-72B (See Tab.~\ref{tab:qwen2_vl_72B}), achieving $\mathbf{93.9\%}$ with only $11.1\%$ tokens, outperforming all competitors significantly. 
\textbf{(iv)} Inspired by~\citep{wen2025token}, we apply \algname during training. \algname$^\dagger$ in Table~\ref{tab:main_1_5} shows better performance-efficiency trade-offs, maintaining full performance with just 192 visual tokens, highlighting the strong adaptability of our method.
These results demonstrate \algname's efficiency in leveraging limited tokens while preserving critical information, showcasing robust performance across tasks, model architectures, and model size.
For more comparisons, please refer to Tables~\ref{tab:qwen2vl},~\ref{tab:minicpm}, and Appendix~\ref{app_sec:more_experimental_results}.
\\
\textbf{Video Understanding Task.}  
To assess \algname's capabilities in video understanding, we integrate it with Video-LLaVA~\citep{lin2023video} and benchmark it against state-of-the-art methods, including FastV~\citep{chen2024image}. Following established protocols, Video-LLaVA processes videos by sampling $8$ frames and extracting $2048$ vision tokens, with $50\%$ retained for evaluation. As demonstrated in Table~\ref{tab:main_table_video}, \algname surpasses FastV across all benchmarks, achieving a notable $4.0$ score on MSVD, $46.3\%$ accuracy on TGIF, and $56.7\%$ accuracy on MSRVT. With an average accuracy of $58.0\%$ and an evaluation score of $3.7$, \algname 
demonstrates superior reasoning over complex multimodal data.

\begin{table*}[htbp]
    \vspace{-2mm}
    \centering
    \begin{minipage}[t]{0.495\textwidth}
    \resizebox{0.995\textwidth}{!}{\setlength{\tabcolsep}{3pt}
    \renewcommand{\arraystretch}{1.20}
    {\begin{tabular}{p{2.5cm} | c c c c c c c | >{\centering\arraybackslash}p{1.2cm}}
        \toprule[1.5pt]
        \textbf{Method} & \textbf{GQA} & \textbf{MMB} & \textbf{MMB-CN} & \textbf{MME} & \textbf{POPE} & \textbf{SQA} & \textbf{VQA}$^{\text{Text}}$ & \makecell[c]{\textbf{Avg}.}\\
        \hline
        \rowcolor{mygray}
        Qwen2-VL-7B & \multicolumn{8}{c}{\textit{Upper Bound, All Tokens} \ $\textbf{(100\%)}$}   \\
        \textcolor{gray}{Vanilla} & \textcolor{gray}{62.2} & \textcolor{gray}{80.5} & \textcolor{gray}{81.2} & \textcolor{gray}{2317} & \textcolor{gray}{86.1} & \textcolor{gray}{84.7} & \textcolor{gray}{82.1} & \multirow{1}*{\textcolor{gray}{100\%}} \\
        \hline

        \rowcolor{mygray}
        Qwen2-VL-7B & \multicolumn{8}{c}{\textit{Token Reduction} \ $\fg{(\downarrow 66.7\%)}$}   \\
        \hspace{0.2em} + FastV \texttt{\scriptsize{(ECCV24)}} & 58.0 & 76.1 & 75.5 & 2130 & 82.1 & 80.0 & 77.3 & 94.0\%  \\
        \hspace{0.2em} + \algname (Ours) & 60.2 & 78.9 & 78.0 & 2245 & 83.9 & 81.4 & 80.5 & 97.0\%  \\
        \hline

        \rowcolor{mygray}
        Qwen2-VL-7B & \multicolumn{8}{c}{\textit{Token Reduction} \ $\fg{(\downarrow 77.8\%)}$}\\
        \hspace{0.2em} + \multirow{1}*{FastV \texttt{\scriptsize{(ECCV24)}}} & 56.7 & 74.1 & 73.9 & 2031 & 79.2 & 78.3 & 72.0 & 91.0\% \\
        \hspace{0.2em} + \algname (Ours) & 58.5 & 77.3 & 77.1 & 2175 & 82.1 & 79.6 & 75.3 & 94.3\%  \\
        \hline

        \rowcolor{mygray}
        Qwen2-VL-7B & \multicolumn{8}{c}{\textit{Token Reduction} \ $\fg{(\downarrow 88.9\%)}$}\\
        \hspace{0.2em} + \multirow{1}*{FastV \texttt{\scriptsize{(ECCV24)}}} & 51.9 & 70.1 & 65.2 & 1962 & 76.1 & 75.8 & 60.3 & 84.0\%  \\
        \hspace{0.2em} + \algname (Ours) & 55.5 & 72.0 & 71.7 & 2052 & 77.9 & 77.6 & 61.8 & 87.5\%  \\

        \bottomrule[1.5pt]
	\end{tabular}}}
        \vspace{-2mm}
	\caption{Comparative Experiments on Qwen2-VL-7B.}
    \vspace{-3mm}
    \label{tab:qwen2vl}
 
    \end{minipage}
    \hfill 
    \begin{minipage}[t]{0.495\textwidth}
\resizebox{0.995\textwidth}{!}{\setlength{\tabcolsep}{3pt}
\renewcommand{\arraystretch}{1.20}
{\begin{tabular}{p{2.5cm} | c c c c c c c | >{\centering\arraybackslash}p{1.2cm}}
        \toprule[1.5pt]
        \textbf{Method} & \textbf{GQA} & \textbf{MMB} & \textbf{MMB-CN} & \textbf{MME} & \textbf{POPE} & \textbf{SQA} & \textbf{VQA}$^{\text{Text}}$ & \makecell[c]{\textbf{Avg}.}\\
        \hline
        \rowcolor{mygray}
        MiniCPM-V2.6 & \multicolumn{8}{c}{\textit{Upper Bound, All Tokens} \ $\textbf{(100\%)}$}   \\
        \textcolor{gray}{Vanilla} & \textcolor{gray}{51.5} & \textcolor{gray}{79.7} & \textcolor{gray}{77.9} & \textcolor{gray}{2267} & \textcolor{gray}{83.2} & \textcolor{gray}{95.6} & \textcolor{gray}{78.5} & \multirow{1}*{\textcolor{gray}{100\%}} \\
        \hline

        \rowcolor{mygray}
        MiniCPM-V2.6 & \multicolumn{8}{c}{\textit{Token Reduction} \ $\fg{(\downarrow 66.7\%)}$}   \\
        \hspace{0.2em} + FastV \texttt{\scriptsize{(ECCV24)}} & 43.2 & 74.9 & 73.1 & 1895 & 75.4 & 89.8 & 67.1 & 89.0\%  \\
        \hspace{0.2em} + \algname (Ours) & 47.8 & 76.5 & 74.8 & 1951 & 77.4 & 91.8 & 70.9 & 92.9\%  \\
        \hline

        \rowcolor{mygray}
        MiniCPM-V2.6 & \multicolumn{8}{c}{\textit{Token Reduction} \ $\fg{(\downarrow 77.8\%)}$}\\
        \hspace{0.2em} + FastV \texttt{\scriptsize{(ECCV24)}} & 41.3 & 72.9 & 70.4 & 1807 & 70.2 & 86.5 & 54.9 & 83.4\%  \\
        \hspace{0.2em} + \algname (Ours) & 47.8 & 73.8 & 71.4 & 1821 & 71.6 & 88.9 & 65.7 & 88.6\%  \\
        \hline

        \rowcolor{mygray}
        MiniCPM-V2.6 & \multicolumn{8}{c}{\textit{Token Reduction} \ $\fg{(\downarrow 88.9\%)}$}  \\
        \hspace{0.2em} + FastV \texttt{\scriptsize{(ECCV24)}} & 35.5 & 61.4 & 60.8 & 1376 & 56.9 & 80.4 & 33.4 & 68.4\%  \\
        \hspace{0.2em} + \algname (Ours) & 42.5 & 66.2 & 64.0 & 1405 & 58.0 & 83.5 & 51.9 & 76.1\%  \\

        \bottomrule[1.5pt]
	\end{tabular}}}
        \vspace{-2mm}
        \caption{Comparative Experiments on MiniCPM-V2.6.}
        \vspace{-4mm}
        \label{tab:minicpm}
    \end{minipage}
\end{table*}

\begin{table}[!ht]
\centering
\scalebox{0.58}{
\setlength{\tabcolsep}{1.2pt}
\arrayrulecolor{black}

\begin{tabular}{@{}l|cc|cc|cc|cc@{}}
\toprule[1.5pt]
\multirow{2}{*}{Methods}       & \multicolumn{2}{c|}{\textbf{TGIF}}& \multicolumn{2}{c|}{\textbf{MSVD}} & \multicolumn{2}{c|}{\textbf{MSRVT}}  & \multicolumn{2}{c}{\textbf{Avg.}}  \\ 
      & Accuracy    &Score             
& Accuracy    & Score        & Accuracy     & Score         & Accuracy    & Score             \\ \hline
FrozenBiLM-1B         & 41.9         &-                 
& 32.2        & -            & 16.8         & -             &30.3 &- \\
VideoChat-7B         & 34.4            &2.3              
& 56.3        & 2.8          & 45.0         & 2.5           &45.1 &2.5 \\
LLaMA-Adapter-7B         & -         &-               
& 54.9        & 3.1          & 43.8         & 2.7           &- & -
\\
Video-LLaMA-7B         & -         &-              
& 51.6        & 2.5          & 29.6         & 1.8           & -& -\\
Video-ChatGPT-7B         & 51.4         &3.0               
& 64.9        & 3.3          & 49.3         & 2.8           &55.2 & 3.0
\\
\hline
\hline
 
Video-LLaVA-7B         & 47.0 &3.4
& 70.2 & 3.9 & 57.3 & 3.5  & 58.2&3.6 \\
\hspace{0.5em} + FastV-7B         &45.2 &3.1
& \textbf{71.0} &3.9 &  55  & 3.5   & 57.1 &3.5 \\

\rowcolor{cyan!10}
\hspace{0.5em} + \algname-7B (Ours)     &\textbf{46.3} &\textbf{3.4}&\textbf{71.0} &\textbf{4.0} & \textbf{56.7}  &\textbf{3.6} & \textbf{58.0} & \textbf{3.7} \\ 

\bottomrule[1.5pt]

\end{tabular}}
\vspace{-2mm}
\caption{\textbf{Comparing MLLMs on Video Understanding tasks} with 50\% visual tokens retained. 
} 
\vspace{-6mm}
\label{tab:main_table_video}
\end{table}


\vspace{-2mm}
\section{Analysis and Discussion}\label{sec:discussion}
\vspace{-1mm}
\subsection{Efficiency Analysis}
\vspace{-1mm}
As shown in Table~\ref{tab:efficiency}, we compare the total inference time, prefill time, FLOPs, and KV cache memory of multiple methods.
\textbf{(i)}  DART achieves a \textbf{2.99$\times$ speedup} in prefill and \textbf{1.99$\times$ speedup} in inference, while its performance on POPE degrades by less than \textbf{3\%} versus the vanilla model.
\textbf{(ii)} Analysis reveals \emph{{although FLOPs reduction is similar across methods, their speeds vary significantly.}} For instance, SparseVLM increases FLOPs by \textbf{2.8\%} versus DART, but its speedup drops \textbf{21.6\%}, showing FLOPs alone poorly measure acceleration.
\textbf{(iii)} We evaluate performance-latency trade-offs using actual latency. Figure~\ref{fig:latency_vs_performance} shows \emph{{some methods underperform random token retention}}. SparseVLM and MustDrop suffer speed degradation from sequential token processing.
FastV's biased attention scores yield worse performance. In contrast, \algname integrates Flash Attention with under \textbf{0.08s} overhead, achieving better performance-speed balance.

\vspace{-2mm}
\subsection{Influence from Selection of Pivot Tokens}\label{pivot_token_selection}
\vspace{-2mm}
In this section, we investigate whether pivot token selection in \algname significantly affects its performance. 
Table~\ref{tab:pivot_token_selection} in Appendix~\ref{app:pivot_selection} evaluates pivot tokens based on criteria such as maximum ($\spadesuit$), minimum ($\heartsuit$) attention scores, K-norm, V-norm, and random selection.
Results show that various strategies achieve over $94.9\%$ of the vanilla model's performance across benchmarks. \ul{\emph{Even DART with randomly selected pivot tokens incurs only a $1.2\%$ performance drop compared to the best strategy and outperforms the previous importance-based methods by $2.1\%$.}}
This observation shows the robustness in the selection of pivot tokens in DART, and 
highlights the crucial role of duplication in token reduction, as \ul{\emph{selecting ``important'' pivot tokens based on attention scores is only 0.2\% better than selecting ``unimportant'' ones as pivot tokens}}.


Furthermore, on the MME benchmark, we analyze the visual tokens retained by selecting pivot tokens based on K-norm$^\spadesuit$ and K-norm$^\heartsuit$. Interestingly, statistical analysis shows that the overlap between tokens preserved by these two strategies is, on average, less than \textbf{50\%}. Despite this low overlap, both strategies achieve highly effective results, {\emph{indicating the existence of multiple distinct groups of tokens which should not be pruned}}. This finding challenges the conventional notion of a single critical token set defined by importance scores, demonstrating that diverse token subsets with minimal overlap can yield comparable performance.

\vspace{-2mm}
\subsection{Influence from Choice of the Pruned Layer and the Number of Pivot Tokens }\label{sec:Layer_and_num}
\vspace{-1mm}

\begin{wrapfigure}{r}{0.53\linewidth}
    \vspace{-3mm}
    \centering
    \includegraphics[width=\linewidth]{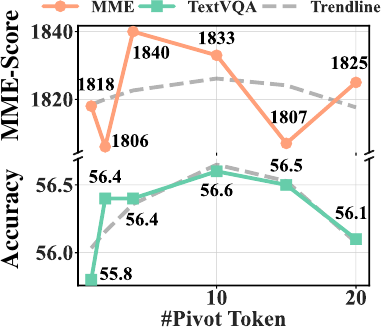}
    \vspace{-7mm}
    \caption{Impact of the number of pivot tokens.}
    \vspace{-4mm}
    \label{fig:sensitivity}
\end{wrapfigure}
\begin{figure}[h]
 \vspace{-2mm}
    \centering
    \subfigure[POPE]{
        \includegraphics[width=0.468\linewidth]{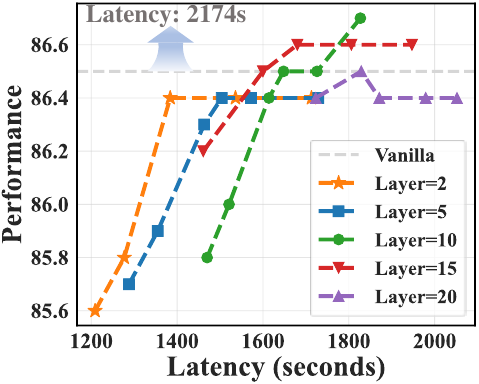}
        \label{fig:pope_layer_scaling_law}
    }
    \subfigure[MME]{
        \includegraphics[width=0.470\linewidth]{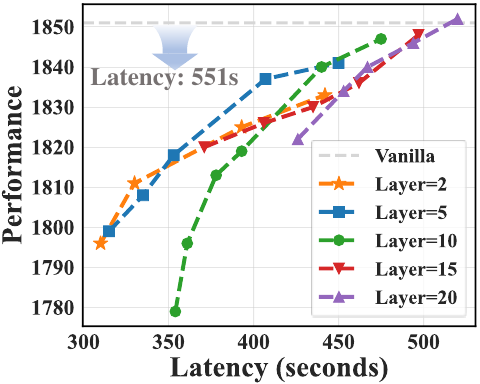}
        \label{fig:mme_layer_scaling_law}
    }
    \vspace{-4mm}
    \caption{Influence from the layer for token pruning.}
    \vspace{-6mm}
    \label{fig:layer_scaling_law}
\end{figure}

We explore the impact of layer on model performance. As expected, pruning deeper layers yields performance closer to the vanilla model but increases latency, as shown in Figure~\ref{fig:layer_scaling_law}. However, we observe two intriguing findings: \textbf{(i)} Pruning at layers 10, 15, and 20 surprisingly outperforms the vanilla model (Fig.~\ref{fig:pope_layer_scaling_law}), consistent with Fig.~\ref{fig:teaser_curry}, suggesting that removing duplicate tokens may reduce hallucinations in MLLMs on the POPE. \textbf{(ii)} At deeper layers (\emph{e.g.}, 15, 20), the latency-minimizing points correspond to pruning all vision tokens, yet performance drops only by $\mathbf{0.1\%} \mathbf{\sim} \mathbf{1.6\%}$. This highlights a modality imbalance in MLLMs, indicating underutilization of the visual modality.
Furthermore, we delved into the impact of the number of pivot tokens on performance. 
As depicted in Figure~\ref{fig:sensitivity}, choosing either an insufficient or an excessive number of pivot tokens leads to suboptimal outcomes.
When a limited number of pivot tokens (\emph{e.g.}, one or two), the lack of diversity among these tokens may impede their ability to comprehensively represent the entire feature space. In contrast, when an overly large number of pivot tokens, for example, 20 or more, are chosen, the majority of retained visual tokens tend to be pivot tokens. In extreme cases, our approach starts to resemble the importance-based method, where pivot tokens essentially transform into important tokens, overlooking the impact of duplication factors.

\subsection{Influence from Modalities of Pivot Tokens}
\begin{figure}[!h]
    \vspace{-3mm}
    \centering
    \subfigure[MME]{
        \includegraphics[width=0.221\textwidth]{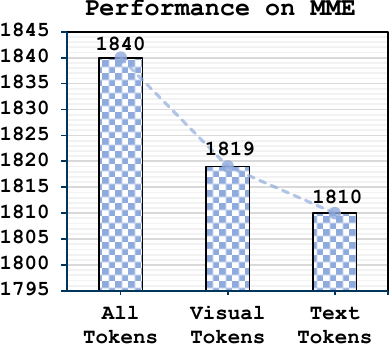}
    }
    \subfigure[TextVQA]{
        \includegraphics[width=0.223\textwidth]{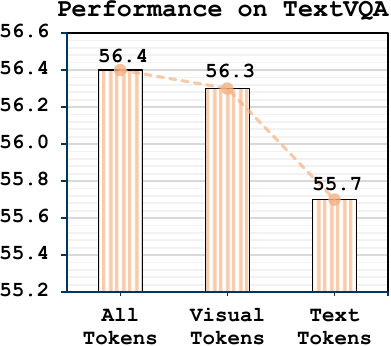}
    }
    \vspace{-3mm}
    \caption{
    \textbf{Analysis of pivot token sources}: ``ALL Tokens'' selects from both visual and textual modalities, while ``Visual Tokens'' and ``Text Tokens'' select exclusively from visual or textual modalities, respectively.}
    \vspace{-5mm}
    
    \label{fig:pivot_tokens}
\end{figure}

We further analyze the impact of the source of pivot tokens on the overall performance of \algname, with a particular focus on understanding whether guidance from the language modality is essential for effective token reduction. 
We evaluate the performance implications of selecting pivot tokens exclusively from either the visual or text modality, aiming to quantify the influence of each modality.
As illustrated in Figure~\ref{fig:pivot_tokens}, the absence of pivot tokens from either modality leads to a noticeable decline in performance. 
This demonstrates that information from both modalities contributes to the token reduction process to varying degrees. Moreover, it highlights that we provide an effective method for incorporating textual guidance without the need to explicitly compute cross-modal attention scores while remaining compatible with Flash Attention.



\section{Conclusion}
The pursuit of efficient token reduction in MLLMs has traditionally focused on token ``importance'', often measured by attention scores, but sometimes performs worse than random pruning. This study introduces \algname, which targets token duplication, removing tokens similar to others and achieving better balance between performance and latency across multiple benchmarks and MLLMs (Tab.~\ref{tab:main_1_5},~\ref{tab:efficiency},~\ref{tab:main_1_6},~\ref{tab:qwen2vl},~\ref{tab:minicpm},~\ref{tab:qwen2_vl_72B},~\ref{app_tab:llava_13B} and Fig.~\ref{fig:latency_vs_performance}). Our exploration yields surprising insights: distinct retained token sets, with under 50\% overlap, deliver similarly strong performance (\secref{pivot_token_selection}). Moreover, token pruning may reduce hallucinations (\secref{sec:Layer_and_num}). These findings expose limits of importance-based methods and offer insights into vision tokens in MLLMs.

\clearpage
\section{Limitations}
Similar to many other methods aimed at improving efficiency, such as network pruning, quantization, distillation, model merging, and speculative decoding, one of the limitations of our work is that it cannot be applied to black-box models like the GPT (\emph{e.g.} GPT 3.5 and more advanced versions) and Claude series, as we are unable to access their encoded tokens during the inference process. 
Moreover, due to space limitations in the main text, we had to move some experimental results that we believe are particularly insightful and interesting to the appendix. These include, for example, our investigation of strategies for pivot token selection, a more detailed analysis of the impact of the number of pivot tokens, and validations of our method on larger-scale models, which may slightly affect the overall reading experience.

\bibliography{custom}

\clearpage

\appendix
\startcontents[appendix]
\printcontents[appendix]{ }{0}{\section*{Appendix}}

\section{Additional Experiments}
\subsection{Supplementary Results on Pivot Token Selection}\label{app:pivot_selection}
This section presents comprehensive experimental results conducted on the LLaVA-1.5-7B model, supporting the analysis of pivot token selection strategies within \algname. Table~\ref{tab:pivot_token_selection} details performance metrics across multiple benchmarks, including GQA, MMB, MME, POPE, SQA, and VQA, with all experiments retaining 128 vision tokens. These findings further validate the robustness of \algname under various pivot token selection criteria, ranging from random selection to methods based on attention scores and norm-based approaches. The table also includes comparisons with baseline methods (\emph{e.g.}, SparseVLM and FastV), highlighting the consistent superiority of \algname across different configurations. For additional insights, refer to the main discussion in \secref{pivot_token_selection}.

\subsection{Influence from the Number of Pivot Tokens}
\label{app_sec:pivot_token_num}
\begin{figure*}[!th]
    \centering
    \includegraphics[width=1.0\linewidth]{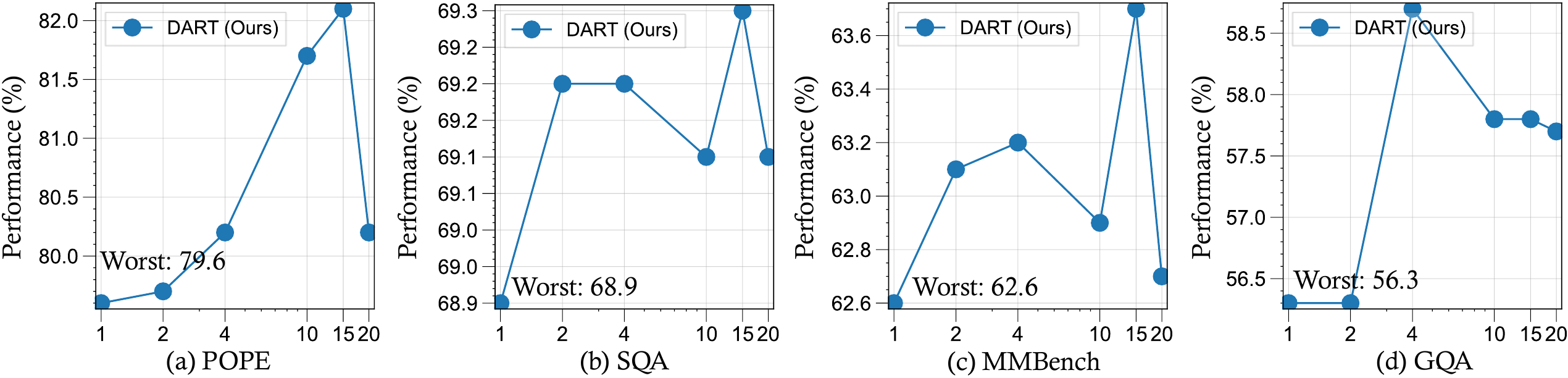}
    \caption{Impact of the number of pivot tokens on performance across additional visual benchmarks. All experiments are conducted with a token reduction ratio of 77.8\%. It is noteworthy that even under relatively extreme numbers of pivot tokens, our worst-case performance still matches or surpasses that of existing token pruning methods.}
    \label{fig:more_pivot_token_num}
    \vspace{-4mm}
\end{figure*}

Beyond the investigation of pivot token numbers on MME and TextVQA in \secref{sec:Layer_and_num}, we conduct additional experiments on several representative visual benchmarks to further support our insight. Figure~\ref{fig:more_pivot_token_num} illustrates that our observations on benchmarks such as POPE and SQA align with those in \secref{sec:Layer_and_num}—namely, that both insufficient and excessive pivot tokens can lead to suboptimal performance. While an insufficient or excessive number of pivot tokens may result in suboptimal outcomes, our statistical analysis reveals that \textbf{even the worst-performing settings still match or surpass the performance of existing token pruning approaches.} This further demonstrates the superiority of \algname.

\subsection{More Experimental Results on Larger MLLMs}
\label{app_sec:more_experimental_results}
\renewcommand{\multirowsetup}{\centering}
\definecolor{mygray}{gray}{.92}
\definecolor{ForestGreen}{RGB}{34,139,34}
\definecolor{Forestred}{RGB}{220,50,50}
\begin{table}[h]
    \centering
    \setlength{\tabcolsep}{3pt}
    \renewcommand{\arraystretch}{1.25}
    \footnotesize
    \centering
    \scalebox{0.85}{
    \begin{tabular}{c | c c c c c | >{\centering\arraybackslash}p{1.0cm}}
        \toprule[1.5pt]
        \textbf{Method} & \textbf{MME} & \textbf{POPE} & \textbf{GQA} & \textbf{TextVQA} & \textbf{SQA} & \makecell[c]{\textbf{Avg}.} \\
        \hline
        \rowcolor{mygray}
        Qwen2-VL-72B & \multicolumn{6}{c}{\textit{Upper Bound, Full Tokens} \ $\textbf{(100\%)}$}\\
        \textcolor{gray}{Vanilla} & \textcolor{gray}{2521} & \textcolor{gray}{87.4} & \textcolor{gray}{65.3} & \textcolor{gray}{82.8} & \textcolor{gray}{91.6} & \textcolor{gray}{100\%} \\
        \hline
        \rowcolor{mygray}
        Qwen2-VL-72B & \multicolumn{6}{c}{\textit{Token Reduction} \ $\fg{(\downarrow 66.7\%)}$}\\ 
        FastV \texttt{\scriptsize{(ECCV24)}} & 2376 & 83.8 & 62.5 & 81.5 & 87.6 & 96.0\% \\
        DART (Ours) & 2511 & 85.7 & 64.2 & 82.1 & 90.9 & 98.9\% \\
        \hline
        \rowcolor{mygray}
        Qwen2-VL-72B & \multicolumn{6}{c}{\textit{Token Reduction} \ $\fg{(\downarrow 77.8\%)}$}\\ 
        FastV \texttt{\scriptsize{(ECCV24)}} & 2219 & 81.1 & 59.2 & 79.6 & 85.1 & 92.1\% \\
        DART (Ours) & 2496 & 83.8 & 62.5 & 80.4 & 88.1 & 96.8\% \\
        \hline
        \rowcolor{mygray}
        Qwen2-VL-72B & \multicolumn{6}{c}{\textit{Token Reduction} \ $\fg{(\downarrow 88.9\%)}$}\\
        FastV \texttt{\scriptsize{(ECCV24)}} & 2089 & 78.7 & 55.7 & 75.4 & 83.3 & 88.0\% \\
        DART (Ours) & 2350 & 79.3 & 59.2 & 76.6 & 86.0 & 92.2\% \\
        \bottomrule[1.5pt]
    \end{tabular}}
    \caption{Comparative experiments on Qwen2-VL-72B.}
    \label{tab:qwen2_vl_72B}
\end{table}

\begin{table*}[!h] 
    \centering
    \scalebox{0.95}{
    \setlength{\tabcolsep}{2.2pt} 
    \renewcommand{\arraystretch}{1.15} 
    \footnotesize
    \begin{tabular}{c |c | c  c  c c c c c | c c}
    \toprule[1.5pt]
    \centering
    \multirow{2}{*}{\textbf{Benchmark}} & \multirow{2}{*}{\textcolor{gray}{\textbf{Vanilla}}} & \multicolumn{7}{c|}{\textbf{Pivot Token Selection}} &  \multicolumn{2}{c}{\textbf{Other Methods}} \\
    \cline{3-11}
    ~ && \cellcolor{mygray}\textbf{Random} & \cellcolor{mygray}\textbf{A-Score}$^\spadesuit$ & \cellcolor{mygray}\textbf{A-Score}$^\heartsuit$ & \cellcolor{mygray}\textbf{K-norm}$^\spadesuit$ & \cellcolor{mygray}\textbf{K-norm}$^\heartsuit$ & \cellcolor{mygray}\textbf{V-norm}$^\spadesuit$ & \cellcolor{mygray}\textbf{V-norm}$^\heartsuit$ & \textbf{SparseVLM} & \textbf{FastV} \\
    \hline
    \textbf{GQA} & \textcolor{gray}{$61.9$} & $59.0_{\pm 0.3}$ & $59.2$ & $58.4$ & $58.7$ & $59.1$ & $57.3$ & $59.4$ & $56.0$ & $49.6$ \\
    
    \textbf{MMB} & \textcolor{gray}{$64.7$} & $63.2_{\pm 0.7}$ & $63.1$ & $62.9$ & $63.2$ & $64.0$ & $62.5$ & $64.3$ & $60.0$ & $56.1$ \\

    \textbf{MME} & \textcolor{gray}{$1862$} & $1772_{\pm 17.9}$ & $1826$ & $1830$ & $1840$ & $1820$ & $1760$ & $1825$ & $1745$ & $1490$ \\

    \textbf{POPE} & \textcolor{gray}{$85.9$} & $80.6_{\pm 0.49}$ & $81.1$ & $81.0$ & $80.1$ & $80.2$ & $76.8$ & $81.6$ & $80.5$ & $59.6$ \\
    
    \textbf{SQA} & \textcolor{gray}{$69.5$} & $69.0_{\pm 0.3}$ & $69.9$ & $68.9$ & $69.1$ & $68.7$ & $69.2$ & $68.9$ & $68.5$ & $60.2$ \\

    \textbf{VQA}$^{\text{V2}}$ & \textcolor{gray}{$78.5$} & $75.2_{\pm 0.2}$ & $75.9$ & $76.0$ & $75.9$ & $75.6$ & $75.4$ & $76.1$ & $73.8$ & $61.8$ \\

    \textbf{VQA}$^{\text{Text}}$ & \textcolor{gray}{$58.2$} & $56.0_{\pm 0.3}$ & $55.7$ & $56.5$ & $56.4$ & $55.4$ & $55.5$ & $56.0$ & $54.9$ & $50.6$ \\

    \hline
    \textbf{Avg.} & \textcolor{gray}{$100\%$} & $96.0\%$ & $96.9\%$ & $96.7\%$ & $96.8\%$ & $96.8\%$ & $94.9\%$ & $97.2\%$ & $93.9\%$  & $81.5\%$ \\
    \bottomrule[1.5pt]
    \end{tabular}}
    \caption{\textbf{Analysis on how to select the pivot token.} 
    This study evaluates pivot tokens, comprising a fixed set of $4$ visual and $4$ text tokens, using various criteria with 128 retained tokens. 
    \textbf{A-Score} denotes the Attention Score. $\spadesuit$ represents selecting token with the highest value as the pivot token. $\heartsuit$ represents selecting the token with the smallest value as the pivot token. For instance, \textbf{A-Score}$^\spadesuit$ means selecting the token with the highest value of Attention Score as the pivot token. For the \textbf{Random} pivot token selection strategy, we conducted experiments five times using five different random seeds, and report the corresponding standard deviation to reflect variability.}
    \label{tab:pivot_token_selection}
\end{table*}

While prior experiments primarily focused on models with 7B parameters, we further validate the effectiveness and robustness of DART on substantially larger models, including LLaVA-v1.5-13B\footnote{\url{https://huggingface.co/liuhaotian/llava-v1.5-13b}} and Qwen2-VL-72B\footnote{\url{https://huggingface.co/Qwen/Qwen2-VL-72B-Instruct}}. Our results demonstrate that DART consistently outperforms prior token pruning methods such as FastV~\citep{chen2024image} and SparseVLM~\citep{zhang2024sparsevlm} across various pruning ratios and downstream tasks, while maintaining near-Vanilla performance.


As shown in Table~\ref{app_tab:llava_13B}, on LLaVA-1.5-13B with an 88.9\% pruning ratio, \algname achieves 94.7\% average performance, significantly outperforming SparseVLM (79.7\%) and FastV (81.0\%). Similarly, on Qwen2-VL-72B, \algname reaches 92.2\% under the same pruning ratio, surpassing FastV (88.0\%) (Table~\ref{tab:qwen2_vl_72B}). At a moderate 66.7\% pruning ratio, \algname retains 99.5\% and 98.9\% accuracy on LLaVA-1.5-13B and Qwen2-VL-72B, respectively, with minimal degradation.

\algname also excels on specific tasks, achieving 60.9 GQA on LLaVA-1.5-13B at 77.8\% pruning and 90.9 ScienceQA on Qwen2-VL-72B at 66.7\%, both outperforming FastV. These results demonstrate \algname’s scalability and its ability to balance compression and performance in large MLLMs.
\renewcommand{\multirowsetup}{\centering}
\definecolor{mygray}{gray}{.92}
\definecolor{ForestGreen}{RGB}{34,139,34}
\definecolor{Forestred}{RGB}{220,50,50}
\begin{table*}[!h]
    \centering
    \setlength{\tabcolsep}{3.5pt}
    \renewcommand{\arraystretch}{1.0}
    \footnotesize
    \centering
    \scalebox{1.0}{
    \begin{tabular}{c | c c c c c c c c | >{\centering\arraybackslash}p{1.0cm}}
        \toprule[1.5pt]
        \textbf{Method} & \textbf{GQA} & \textbf{MMB} & \textbf{MMB-CN} & \textbf{MME} & \textbf{POPE} & \textbf{SQA} &  \textbf{VQA}$^{\text{Text}}$ & \textbf{VizWiz} & \makecell[c]{\textbf{Avg}.}  \\
        \hline
        \rowcolor{mygray}
        LLaVA-1.5-13B & \multicolumn{9}{c}{\textit{Upper Bound, 576 Tokens} \ $\textbf{(100\%)}$}\\
        \textcolor{gray}{Vanilla} & \textcolor{gray}{63.3} & \textcolor{gray}{68.9} & \textcolor{gray}{62.3} & \textcolor{gray}{1818} & \textcolor{gray}{85.9} & \textcolor{gray}{72.8} & \textcolor{gray}{61.3} &  \textcolor{gray}{56.6} & \multirow{1}*{\textcolor{gray}{100\%}} \\
        \hline

        \rowcolor{mygray}
        LLaVA-1.5-13B & \multicolumn{9}{c}{\textit{Retain 192 Tokens} \ $\fg{(\downarrow 66.7\%)}$}\\
        FastV \texttt{\scriptsize{(ECCV24)}} & 59.1 & 54.0 & 51.2 & 1641 & 82.3 & 56.4 & 51.6 & 56.9 & 87.8\% \\
        SparseVLM \texttt{\scriptsize{(ICML25)}} & 58.7 & 67.4 & 61.0 & 1768 & 82.2 & 73.1 & 45.4 & 56.5 & 94.5\% \\
        \algname (Ours) & \textbf{62.1} & \textbf{68.2} & \textbf{61.4} & \textbf{1855} & \textbf{84.0} & \textbf{73.6} & \textbf{60.2} & \textbf{57.3} & \textbf{99.5\%} \\
        \hline

        \rowcolor{mygray}
        LLaVA-1.5-13B & \multicolumn{9}{c}{\textit{Retain 128 Tokens} \ $\fg{(\downarrow 77.8\%)}$}   \\
        FastV \texttt{\scriptsize{(ECCV24)}} & 57.7 & 57.9 & 48.8 & 1673 & 79.3 & 57.0 & 56.0 & 55.3 & 88.2\%  \\
        SparseVLM \texttt{\scriptsize{(ICML25)}} & 57.9 & 65.8 & 55.8 & 1774 & 81.1 & 69.9 & 49.9 & 56.3 & 93.2\% \\
        \algname (Ours) & \textbf{60.9} & \textbf{67.4} & \textbf{60.7} & \textbf{1839} & \textbf{81.8} & \textbf{74.3} & \textbf{59.0} & \textbf{57.3} & \textbf{98.5\%}   \\
        \hline

        \rowcolor{mygray}
        LLaVA-1.5-13B & \multicolumn{9}{c}{\textit{Retain 64 Tokens} \ $\fg{(\downarrow 88.9\%)}$}\\
        FastV \texttt{\scriptsize{(ECCV24)}} & 53.7 & 50.9 & 42.1 & 1567 & 69.3 & 56.8 & 47.1 & 56.7 & 81.0\% \\
        SparseVLM \texttt{\scriptsize{(ICML25)}} & 50.6 & 61.3 & 54.8 & 1402 & 65.0 & 69.0 & 22.7 & 54.5 & 79.7\% \\
        \algname (Ours) & \textbf{57.1} & \textbf{65.4} & \textbf{59.3} & \textbf{1722} & \textbf{75.4} & \textbf{74.1} & \textbf{55.9} & \textbf{57.4} & \textbf{94.7\%} \\

        \bottomrule[1.5pt]
	\end{tabular}}
        \vspace{-2mm}
	\caption{Comparative experiments on LLaVA-1.5-13B. In all experiments for \algname, tokens are pruned after the second layer with $8$ pivot tokens. The pivot tokens are selected based on the maximum K-norm.}
    \label{app_tab:llava_13B}
    \vspace{-3mm}
\end{table*}

\section{Extensions to Other Scenarios}
\label{app_sec:extensions}
\subsection{Exploring the Effectiveness of DART in Audio Modalities}
\renewcommand{\multirowsetup}{\centering}
\definecolor{mygray}{gray}{.92}
\definecolor{ForestGreen}{RGB}{34,139,34}
\definecolor{Forestred}{RGB}{220,50,50}
\begin{table}[!h]
    \centering
    \setlength{\tabcolsep}{3.5pt}
    \renewcommand{\arraystretch}{1.0}
    \footnotesize
    \centering
    \scalebox{0.8}{
    \begin{tabular}{p{3.4cm} | c c  c}
        \toprule[1.5pt]
        \textbf{Method} & \textbf{FLEURs} $\downarrow$ & \textbf{LibriSpeech} $\downarrow$ & \textbf{Avg.} $\downarrow$  \\
        \hline
        \rowcolor{mygray}
        Phi-4-Multimodal-Instruct & \multicolumn{3}{c}{\textit{Upper Bound, Full Audio Tokens} \ $\textbf{(100\%)}$}\\
        \hspace{1em} \textcolor{gray}{Vanilla} & 3.49 & 6.40 & 4.95 \\
        \hline

        \rowcolor{mygray}
        Phi-4-Multimodal-Instruct & \multicolumn{3}{c}{\textit{Token Reduction} \ $\fg{(\downarrow 20\%)}$}\\
        \hspace{0.8em} + Random & 8.15 & 25.23 & 16.69 \\
        \hspace{0.8em} + FastV \texttt{\scriptsize{(ECCV24)}} & 19.82 & 27.90 & 23.86 \\
        \hspace{0.8em} + \algname (Ours) & \textbf{5.05} & \textbf{6.95} & \textbf{6.00} \\
        \hline

        \rowcolor{mygray}
        Phi-4-Multimodal-Instruct & \multicolumn{3}{c}{\textit{Token Reduction} \ $\fg{(\downarrow 30\%)}$}   \\
        \hspace{0.8em} + Random & 13.18 & 39.42 & 26.3 \\
        \hspace{0.8em} + FastV \texttt{\scriptsize{(ECCV24)}} & 34.10 & 51.60 & 42.85 \\
        \hspace{0.8em} + \algname (Ours) & \textbf{5.84} & \textbf{11.64} & \textbf{8.74} \\
        \hline

        \rowcolor{mygray}
        Phi-4-Multimodal-Instruct & \multicolumn{3}{c}{\textit{Token Reduction} \ $\fg{(\downarrow 50\%)}$}\\
        \hspace{0.8em} + Random & 37.57 & 76.85 & 57.21 \\
        \hspace{0.8em} + FastV \texttt{\scriptsize{(ECCV24)}} & 180.0 & 88.38 & 134.19 \\
        \hspace{0.8em} + \algname (Ours) & \textbf{18.93} & \textbf{49.13} & \textbf{34.03} \\

        \bottomrule[1.5pt]
	\end{tabular}}
	\caption{Comparative experiments on Automatic Speech Recognition tasks. In all experiments for \algname, tokens are pruned after the 2nd layer with $8$ pivot tokens. The pivot tokens are selected based on the maximum K-norm. The evaluation metric is Word Error Rate (WER).}
    \label{app_tab:audio}
    \vspace{-4mm}
\end{table}
In recent years, the integration of audio as a core modality~\citep{abouelenin2025phi,kimi_audio_2024,Qwen2-Audio} within Multimodal Large Language Models (MLLMs) has garnered increasing attention. As these models evolve to handle complex, real-world tasks that span language, vision, and sound, the ability to effectively process spoken language becomes crucial. Audio understanding, particularly in the form of automatic speech recognition (ASR), plays a foundational role in applications such as virtual assistants, transcription services, voice-controlled systems, and multimodal reasoning agents.
Therefore, beyond the widely explored domains of image and video understanding in the visual modality, we further extend our investigation to evaluate the effectiveness of our proposed method on tasks within the audio modality.
To conduct our study, we select Phi-4-Multimodal-Instruct\footnote{\url{https://huggingface.co/microsoft/Phi-4-multimodal-instruct}}, an MLLM with strong audio modality capabilities, and evaluate it on two representative speech benchmarks: FLEURs-en~\citep{conneau2023fleurs} and LibriSpeech-long~\citep{park2024long}.
As demonstrated in Table~\ref{app_tab:audio}, our proposed method \algname consistently outperforms baseline approaches under varying token reduction ratios on both FLEURs-en and LibriSpeech-long benchmarks. While random pruning and FastV result in substantial degradation in recognition performance, particularly under higher reduction rates, \algname maintains significantly lower Word Error Rates (WER), showcasing its robustness and effectiveness in preserving critical audio information even with limited token usage.

\subsection{Enhancing VLA Efficiency with \algname}
\definecolor{lightgray}{gray}{.9}
\definecolor{lightblue}{RGB}{230,240,255}
\definecolor{lightgreen}{RGB}{230,255,230}
\definecolor{lightyellow}{RGB}{255,255,230}
\definecolor{lightred}{RGB}{255,230,230}
\definecolor{lightlightgray}{gray}{.95}
\definecolor{lightlightblue}{RGB}{240,245,255}
\definecolor{lightlightgreen}{RGB}{240,255,240}
\definecolor{lightlightyellow}{RGB}{255,255,240}
\definecolor{lightlightred}{RGB}{255,240,240}
\definecolor{lightlightlightgray}{gray}{.99}
\definecolor{lightlightlightblue}{RGB}{247,250,255}
\definecolor{lightlightlightgreen}{RGB}{247,255,247}
\definecolor{lightlightlightyellow}{RGB}{255,255,247}
\definecolor{lightlightlightred}{RGB}{255,247,247}

\begin{table*}[!ht]
\begin{center}
    \setlength\tabcolsep{2.5pt} 
    \renewcommand{\arraystretch}{1.1} 
    \resizebox{1.0\linewidth}{!}{%
        \begin{tabular}{l|c|c|ccccc|cc} 
        \toprule[1.5pt]
        \textbf{SIMPLER} & \textbf{Method} & \textbf{Retained Tokens} & \textbf{PickCan} & \textbf{MoveNear} & \textbf{Drawer} & \textbf{DrawerApple} & \textbf{Average} & \textbf{FLOPs $\downarrow$} & \textbf{Speedup $\uparrow$}  \\ 
        \midrule\midrule 
        \multirow{5}{*}{\begin{tabular}[l]{@{}l@{}}\textbf{Visual} \\ \textbf{Matching}\end{tabular}}  
          & \textcolor{gray}{\textbf{CogACT}} & \textcolor{gray}{256} & \textcolor{gray}{91.3\%} & \textcolor{gray}{85.0\%} & \textcolor{gray}{71.8\%} & \textcolor{gray}{50.9\%} & \textcolor{gray}{74.8\%} & \textcolor{gray}{100.0\%} & \textcolor{gray} {1.00$\times$}  \\
          & \textbf{Random Dropping} & 112 & 9.7\% & 20.4\% & 53.5\% & 0.0\% & 20.9\% & 58.5\% & 1.20$\times$  \\
          & \textbf{FastV} & 56 & 92.6\% & 81.4\% & 69.8\% & 52.4\% & 74.1\% & 42.0\% & 1.21$\times$ \\
          & \textbf{VLA-Cache} & - & 92.0\% & 83.3\% & 70.5\% & 51.6\% & 74.4\% & 80.1\% & 1.38$\times$  \\
          \rowcolor{cyan!7}
          & \textbf{\algname} & 56 & \textbf{95.6\%} & \textbf{85.8\%} & 69.9\% & 49.5\% & \textbf{75.2\%} & 44.7\% & 1.25$\times$ \\
        \midrule
        \multirow{5}{*}{\begin{tabular}[l]{@{}l@{}} \textbf{Variant} \\ \textbf{Aggregation}\end{tabular}}  
          & \textcolor{gray}{\textbf{CogACT}} & \textcolor{gray}{256} & \textcolor{gray}{89.6\%} & \textcolor{gray}{80.8\%} & \textcolor{gray}{28.3\%} & \textcolor{gray}{46.6\%} & \textcolor{gray}{61.3\%} & \textcolor{gray}{100.0\%} & \textcolor{gray}{1.00$\times$}  \\
          & \textbf{Random Dropping} & 112 & 4.0\% & 16.1\% & 15.6\% & 0.0\% & 8.9\% & 58.5\% & 1.20$\times$  \\
          & \textbf{FastV} & 56 & 91.4\% & 78.6\% & 27.6\% & 50.6\% & 62.1\% & 42.0\% & 1.19$\times$ \\
          & \textbf{VLA-Cache} & - & 91.7\% & 79.3\% & 32.5\% & 45.8\% & 62.3\% & 82.6\% & 1.37$\times$  \\
          \rowcolor{cyan!7}
          & \textbf{\algname} & 56 & \textbf{92.4\%} & 77.0\% & \textbf{35.9\%} & \textbf{52.4\%} & \textbf{64.4\%} & 44.7\% & 1.25$\times$ \\
        \bottomrule[1.5pt]
        \end{tabular}
    }
 \caption{Performance of \algname on the CogACT versus the other baselines in the SIMPLER environment. Random Dropping denotes a method involving the random retention of visual tokens.}
 \label{tab:vla_exp}
\end{center}
\vspace{-6mm}
\end{table*}
Building on recent progress in multimodal understanding from vision-language models~\citep{awadalla2023openflamingo, li2022blip, radford2021learning, an2024mc, luo2024llm}, Vision-Language-Action (VLA) models represent a significant step toward embodied intelligence. Systems such as OpenVLA~\citep{kim24openvla}, CogACT~\citep{li2024CogACT}, $pi_0$\citep{black2024pi_0}, and RT-2\citep{brohan2023rt2visionlanguageactionmodelstransfer} seamlessly translate multimodal inputs into executable actions. Leveraging large-scale datasets~\citep{fang2024rh20t, o2024open}, these models have demonstrated impressive capabilities in complex robotic manipulation and reasoning tasks.
As a potential pathway toward Artificial General Intelligence (AGI), we place great emphasis on improving the efficiency of VLA models through our approach.

To this end, we employ the SIMPLER environment~\citep{li2024evaluating}, a simulation-based benchmark specifically designed for table-top manipulation to evaluate our method. SIMPLER aims to closely mirror real-world dynamics observed in robots such as the Google Robot and WidowX, exhibiting strong consistency between simulated and real-world performance.
In this setup, the Vision-Language-Action (VLA) model receives 224$\times$224 RGB image observations along with natural language task instructions (\emph{e.g.}, ``Pick coke can'') and generates a sequence of actions in 7-DoF Cartesian space.
SIMPLER supports two evaluation configurations: \textbf{Visual Matching}, which emphasizes visual fidelity to real-world scenes, and \textbf{Variant Aggregations}, which introduces variability through changes in lighting, background, and surface textures. For the Google Robot, both configurations include the same set of four tasks: Pick coke can; Move near; Open/close drawer and Open top drawer and place apple. Performance is assessed using success rate as the evaluation metric.

As shown in Table~\ref{tab:vla_exp}, \algname demonstrates superior performance compared to other baseline methods in the SIMPLER environment. With only 56 retained visual tokens, \algname achieves the highest average success rates of 75.2\% and 64.4\% in Visual Matching and Variant Aggregation, respectively, outperforming Random Dropping~\citep{wen2025token}, FastV~\citep{chen2024image}, VLA-Cache~\citep{xu2025vla}, and even vanilla CogACT~\citep{li2024CogACT}. Moreover, \algname significantly reduces computational cost, achieving the lower FLOPs (44.7\%), which corresponds to a speedup of 1.25$\times$ compared to the CogACT. These results highlight \algname’s efficiency in maintaining high task performance while substantially reducing computational demands.

\section{Detailed Experiment Settings}\label{app:detailed_experiment_settings}
\subsection{Datasets}
\label{app:dataset}
Our experiments are conducted on a suite of widely recognized benchmarks, each designed to evaluate distinct aspects of multimodal intelligence.
For image understanding task, we performed experiments on ten widely used benchmarks, including GQA \citep{hudson2019gqa}, MMBench (MMB) and MMB-CN \citep{liu2025mmbench}, MME \citep{fu2023mme}, POPE~\citep{li2023evaluating}, VizWiz \citep{bigham2010vizwiz}, SQA \citep{lu2022learn}, VQA$^{\text{V2}}$ (VQA V2) \citep{goyal2017making}, VQA$^{\text{Text}}$ (TextVQA) \citep{singh2019towards}, and OCRBench ~\citep{liu2024ocrbench}.
For video understanding task, we evaluated our method on three video-based benchmarks: TGIF-QA \citep{jang2017tgif}, MSVD-QA \citep{xu2017video}, and MSRVTT-QA \citep{xu2017video}.
Furthermore, to validate the effectiveness and applicability of our approach, we extended the evaluation scenarios of \algname. Specifically, we tested our token reduction method in both the speech modality—on automatic speech recognition (audio token reduction)~\citep{conneau2023fleurs,park2024long}, and on a vision-language-action model within a simulated environment~\citep{li2024evaluating}.

\subsubsection{Image Understanding}
\textbf{GQA.} GQA is structured around three core components: scene graphs, questions, and images. It includes not only the images themselves but also detailed spatial features and object-level attributes. The questions are crafted to assess a model's ability to comprehend visual scenes and perform reasoning tasks based on the image content. \\
\textbf{MMBench.} MMBench offers a hierarchical evaluation framework, categorizing model capabilities into three levels. The first level (L-1) focuses on perception and reasoning. The second level (L-2) expands this to six sub-abilities, while the third level (L-3) further refines these into 20 specific dimensions. This structured approach allows for a nuanced and comprehensive assessment of a model's multifaceted abilities. MMBench-CN is the Chinese version of the dataset. \\
\textbf{MME.} The MME benchmark is designed to rigorously evaluate a model's perceptual and cognitive abilities through 14 subtasks. It employs carefully constructed instruction-answer pairs and concise instructions to minimize data leakage and ensure fair evaluation. This setup provides a robust measure of a model's performance across various tasks. \\
\textbf{POPE.} POPE is tailored to assess object hallucination. It presents a series of binary questions about the presence of objects in images, using accuracy, recall, precision, and F1 score as metrics. This approach offers a precise evaluation of hallucination levels under different sampling strategies. \\
\textbf{ScienceQA.} ScienceQA spans a wide array of domains, including natural, language, and social sciences. Questions are hierarchically categorized into 26 topics, 127 categories, and 379 skills, providing a diverse and comprehensive testbed for evaluating multimodal understanding, multi-step reasoning, and interoperability. \\
\textbf{VQA V2.} VQA V2 challenges models with open-ended questions based on 265,016 images depicting a variety of real-world scenes. Each question is accompanied by 10 human-annotated answers, enabling a thorough assessment of a model's ability to accurately interpret and respond to visual queries. \\
\textbf{TextVQA.} TextVQA emphasizes the integration of textual information within images. It evaluates a model's proficiency in reading and reasoning about text embedded in visual content, requiring both visual and textual comprehension to answer questions accurately. \\
\textbf{VizWiz.} VizWiz is a visual benchmark designed to assist visually impaired individuals. It contains real-world images captured by blind users, paired with questions they ask about the images. The dataset includes 20,523 training, 4,319 validation, and 8,000 test image-question pairs, with each question accompanied by 10 human-annotated answers. VizWiz challenges models to answer questions accurately or determine if a question is answerable, focusing on practical visual understanding and accessibility. \\
\textbf{OCRBench.} OCRBench is a comprehensive benchmark for evaluating the OCR capabilities of multi-modal language models across five key tasks: text recognition, scene text-centric and document-oriented VQA, key information extraction, and handwritten mathematical expression recognition.

\subsubsection{Video Understanding}
\textbf{TGIF-QA.} TGIF-QA extends the image question-answering task to videos, featuring 165,000 question-answer pairs. It introduces tasks that require spatio-temporal reasoning, such as repetition count and state transition, as well as frame-based questions, promoting advancements in video question answering. \\
\textbf{MSVD-QA.} Based on the MSVD dataset, MSVD-QA includes 1970 video clips and approximately 50.5K QA pairs. The questions cover a broad spectrum of topics and are open-ended, categorized into what, who, how, when, and where types, making it a versatile tool for video understanding tasks.
\\
\textbf{MSRVTT-QA.} MSRVTT-QA comprises 10K video clips and 243K QA pairs. It addresses the challenge of integrating visual and temporal information in videos, requiring models to effectively process both to answer questions accurately. Similar to MSVD-QA, it includes five types of questions, further enriching the evaluation landscape.

\subsubsection{Automatic Speech Recognition.}
\textbf{FLEURS.} FLEURS is a benchmark for evaluating universal speech representations across 102 languages, built on top of the FLoRes-101 dataset. It contains 12 hours of speech data per language, with parallel speech and text for tasks like ASR, Speech LangID, and cross-modal retrieval.
\\
\textbf{LibriSpeech-Long.} LibriSpeech-Long is a benchmark dataset for long-form speech generation, derived from the original LibriSpeech dataset. It provides 4-minute long continuous speech and corresponding transcripts, enabling the evaluation of long-form speech continuation. This benchmark supports reference-based evaluation for long-form speech tasks and facilitates research in generating coherent and contextually relevant speech over extended durations.

\subsubsection{Vision-Language-Action Models Simulation Platform}
\textbf{SIMPLER.} SIMPLER is a simulation platform for evaluating real-world robot manipulation policies. It features realistic simulated environments that match common real robot setups (\emph{e.g.}, Google Robot and WidowX) and tasks (\emph{e.g.}, picking and moving objects). By addressing control and visual disparities between simulation and reality, SIMPLER achieves strong correlation with real-world performance, providing a scalable and reproducible evaluation tool.

\subsection{Models}\label{app:models}
We evaluate \algname using various open-source MLLMs. For image understanding tasks, experiments are conducted on the LLaVA family, including LLaVA-1.5-7B\footnote{\url{https://huggingface.co/liuhaotian/llava-v1.5-7b}}~\citep{liu2024visual} and LLaVA-Next-7B\footnote{\url{https://huggingface.co/liuhaotian/llava-v1.6-vicuna-7b}}~\citep{liu2024llavanext}, with the latter used to validate performance on high-resolution images.
Furthermore, we validate our method on more advanced models, including Qwen2-VL-7B\footnote{\url{https://huggingface.co/Qwen/Qwen2-VL-7B-Instruct}}~\citep{wang2024qwen2} and MiniCPM-V-2.6\footnote{\url{https://huggingface.co/openbmb/MiniCPM-V-2_6}}~\citep{yao2024minicpm}.
Moreover, to enhance the effectiveness of our proposed method, we also validate DART on larger MLLMs, such as Qwen2-VL-72B and LLaVA-1.5-13B.
For video understanding tasks, we use Video-LLaVA~\citep{lin2023video} as the baseline model.
following the settings reported in their paper to ensure a fair comparison.



\subsection{Baselines}
We analyze multiple representative methods for accelerating multi-modal language models (MLLMs) through token reduction. These methods share the goal of improving efficiency by reducing redundant tokens, yet differ in their strategies, such as token merging, pruning, or adaptive allocation.
\\
\textbf{ToMe}~\citep{bolya2022tome} merges similar tokens in visual transformer layers through lightweight matching techniques, achieving acceleration without requiring additional training.
\\
\textbf{FastV}~\citep{chen2024image} focuses on early-stage token pruning by leveraging attention maps, effectively reducing computational overhead in the initial layers.
\\
\textbf{SparseVLM}~\citep{zhang2024sparsevlm} ranks token importance using cross-modal attention and introduces adaptive sparsity ratios, complemented by a novel token recycling mechanism.
\\
\textbf{HiRED}~\citep{arif2024hired} allocates token budgets across image partitions based on \texttt{CLS} token attention, followed by the selection of the most informative tokens within each partition, ensuring spatially aware token reduction.
\\
\textbf{LLaVA-PruMerge}~\citep{shang2024llava} combines pruning and merging strategies by dynamically removing less important tokens using sparse \texttt{CLS}-visual attention and clustering retained tokens based on key similarity.
\\
\textbf{PDrop}~\citep{xing2024PyramidDrop} adopts a progressive token-dropping strategy across model stages, forming a pyramid-like token structure that balances efficiency and performance.
\\
\textbf{MustDrop}~\citep{liu2024multi} integrates multiple strategies, including spatial merging, text-guided pruning, and output-aware cache policies, to reduce tokens across various stages.
\\
\textbf{FasterVLM}~\citep{zhang2024cls} evaluates token importance via \texttt{CLS} attention in the encoder and performs pruning before interaction with the language model, streamlining the overall process.
\\
\textbf{GlobalCom$^2$}~\citep{liu2025compression} introduces a hierarchical approach by coordinating thumbnail tokens to allocate retention ratios for high-resolution crops while preserving local details.
\\
\textbf{FiCoCo}~\citep{han2024rethinking} introduces a unified ``filter-correlate-compress'' paradigm to streamline training-free token reduction in Multimodal Large Language Models (MLLMs).
\\
\textbf{FitPrune}~\citep{ye2025fit} proposes a method that generates an efficient token pruning strategy for multi-modal large language models by removing redundant visual tokens. FitPrune is easy to deploy and is designed to meet a predefined computational budget while maintaining model performance.

These methods collectively highlight diverse approaches to token reduction, ranging from attention-based pruning to adaptive merging, offering complementary solutions for accelerating MLLMs.

\subsection{Implementation Details}\label{app:Implementation_details}
All of our experiments are conducted on Nvidia A100-80G GPU. The implementation was carried out in Python 3.10, utilizing PyTorch 2.1.2, and CUDA 11.8. All baseline settings follow the original paper. \\

\section{Computational Complexity.}
To evaluate the computational complexity of MLLMs, it is essential to analyze their core components, including the self-attention mechanism and the feed-forward network (FFN). The total floating-point operations (FLOPs) required can be expressed as:  
\begin{equation}
\text{Total FLOPs} = T \times (4nd^2 + 2n^2d + 2ndm),
\end{equation}  
where $T$ denotes the number of transformer layers, $n$ is the sequence length, $d$ represents the hidden dimension size, and $m$ is the intermediate size of the FFN.  
This equation highlights the significant impact of sequence length $n$ on computational complexity. 
Notable, we follow FastV~\cite{chen2024image} to roughly estimate various token reduction baseline FLOPs.
The FLOPs after token pruning can be represented as:
\begin{equation}
\begin{split}
    \text{Post-Pruning FLOPs} \\
    &\hspace{-7em}= L\times(4nd^2+2n^2d+2ndm) +{} \\
    &\hspace{-5em} (T-L) \times (4\hat{n}d^2+2\hat{n}^2d+2\hat{n}dm),
\end{split}
\end{equation}
where $L$ denotes the pruned layer, $\hat{n}$ represents token sequence length after pruning.
The theoretical FLOPs reduction ratio related to visual tokens is computed as:
\begin{equation}
    1-\frac{\text{Post-Pruning FLOPs}}{\text{Total FLOPs}}.
\end{equation}

\section{Future Works}
As can be observed from Figure~\ref{fig:teaser_curry} and Figure~\ref{fig:pope_layer_scaling_law}, in certain cases, token pruning contributes to the reduction of hallucinations. Our method achieved better results than the vanilla model on the POPE benchmark, which is specifically designed for evaluating the hallucination issues of multimodal large language models. Therefore, we believe that it is worth exploring in the future why token pruning is beneficial for reducing hallucinations and how we can better utilize efficient techniques (\emph{e.g.}, token pruning, and token merge) to reduce hallucinations while achieving acceleration benefits.

\section{Sparsification Visualization on Different Pivot Token Selection Strategy}
\begin{figure*}[!h]
    \centering
    \includegraphics[width=\linewidth]{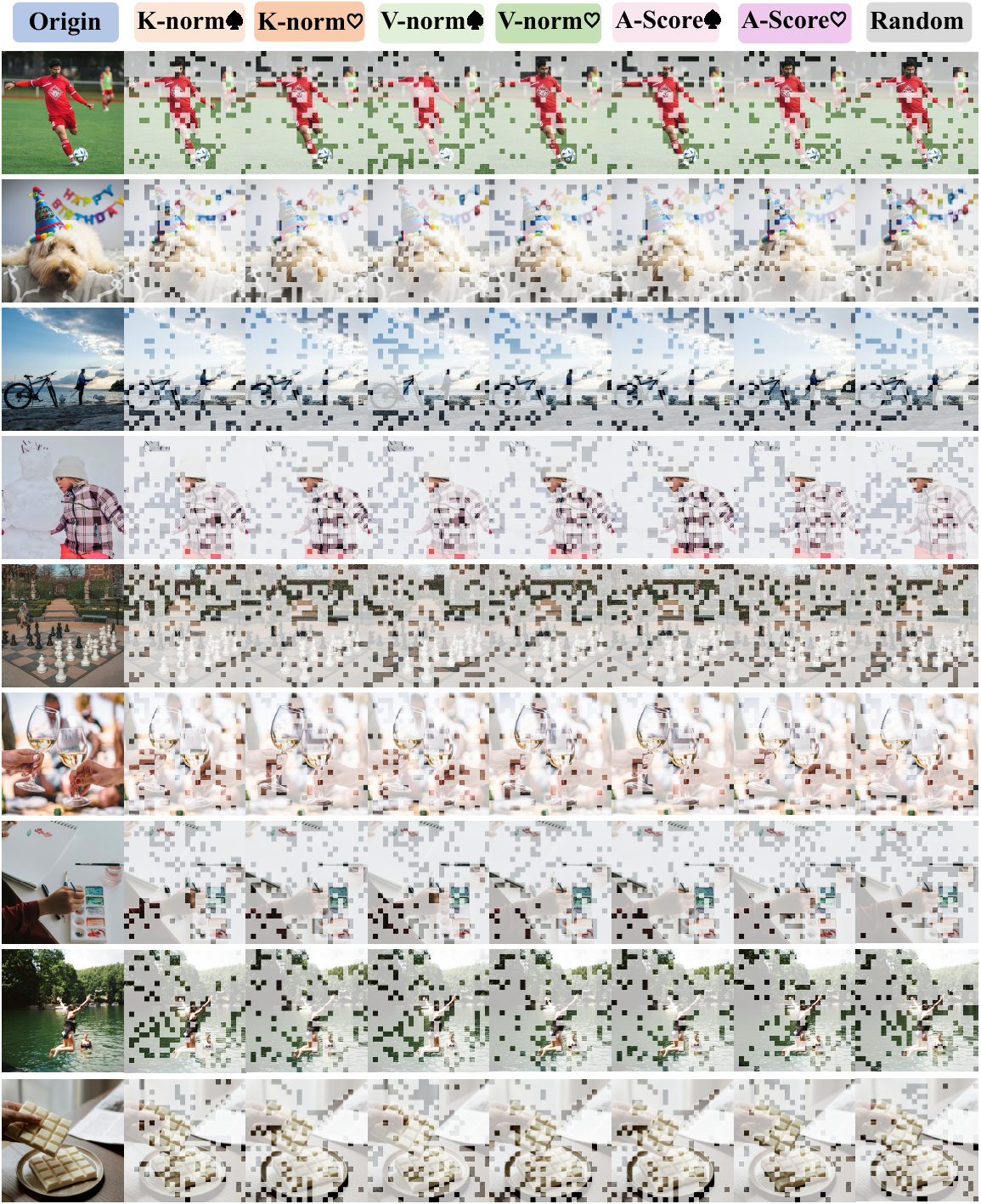}
    \caption{\textbf{Sparsification Visualization examples of DART on different Pivot Token Selection Strategy.}}
    \label{fig:dart_visual_cases}
\end{figure*}
Figure~\ref{fig:dart_visual_cases} showcases a diverse array of sparsification visualization examples on different pivot token selection strategy, including K-norm$\spadesuit$, K-norm$\heartsuit$, V-norm$\spadesuit$, V-norm$\heartsuit$, Attention Score$\spadesuit$, Attention Score$\heartsuit$, and Random.
Here, we can observe two interesting points: (i) The commonality is that \algname employs different pivot token selection strategies for token reduction, and the retained tokens are distributed in a relatively scattered manner without obvious bias, \emph{i.e.}, spatial uniformity, which contributes to a more accurate understanding of the entire image and consistent responses. (ii) The difference lies in the fact that although each strategy achieves comparable performance, it is noticeable that the final set of retained tokens varies significantly across strategies, indicating the existence of multiple token sets that can deliver satisfactory results. This further corroborates the limitation of selecting a unique set of tokens based solely on importance scores.

\end{document}